\documentclass{article} 
\usepackage{iclr2024_conference,times}

\usepackage[utf8]{inputenc} 
\usepackage[T1]{fontenc}    
\usepackage{hyperref}       
\usepackage{url}            
\usepackage{booktabs}       
\usepackage{amsfonts}       
\usepackage{nicefrac}       
\usepackage{microtype}      
\usepackage{graphicx}
\usepackage{subfigure}
\usepackage{booktabs} 
\usepackage{xcolor}         
\usepackage{amsmath, amsthm, amsfonts}

\usepackage{algorithmic, algorithm}
\newtheorem{prop}{Proposition}
\newtheorem{thrm}{Theorem}

\newtheorem{lemma}{Lemma}

\usepackage{verbatim}

\usepackage{amsmath,amsfonts,bm}

\def\eqref#1{equation~\ref{#1}}

\def\1{\bm{1}}

\DeclareMathAlphabet{\mathsfit}{\encodingdefault}{\sfdefault}{m}{sl}
\SetMathAlphabet{\mathsfit}{bold}{\encodingdefault}{\sfdefault}{bx}{n}

\newcommand{\E}{\mathbb{E}}

\newcommand{\R}{\mathbb{R}}

\DeclareMathOperator{\Tr}{Tr}

\title{How connectivity structure shapes rich and lazy learning in neural circuits}

\usepackage{authblk}

\author[1,2,3,*]{Yuhan Helena Liu}
\author[4]{Aristide Baratin}
\author[3,5]{Jonathan Cornford}
\author[1,2]{Stefan Mihalas}
\author[1,2]{Eric Shea-Brown}
\author[3,6,7,*]{Guillaume Lajoie}
\affil[1]{University of Washington, Seattle, WA, USA}
\affil[2]{Allen Institute for Brain Science, Seattle WA, USA}
\affil[3]{Mila - Quebec AI Institute, Montreal, QC, Canada}
\affil[4]{Samsung - SAIT AI Lab, Montreal, QC, Canada}
\affil[5]{McGill University, Montreal, QC, Canada}
\affil[6]{Canada CIFAR AI Chair, CIFAR, Toronto, ON, Canada}
\affil[7]{Université de Montréal, Montreal, QC, Canada}
\affil[*]{Correspondence: hyliu24@uw.edu, g.lajoie@umontreal.ca}

\iclrfinalcopy 

\begin{document}

\maketitle

\begin{abstract}

In theoretical neuroscience, recent work leverages deep learning tools to explore how some network attributes critically influence its learning dynamics. Notably, initial weight distributions with small (resp. large) variance may yield a rich (resp. lazy) regime, where significant (resp. minor) changes to network states and representation are observed over the course of learning. However, in biology, neural circuit connectivity could exhibit a low-rank structure and therefore differs markedly from the random initializations generally used for these studies. As such, here we investigate how the structure of the initial weights --- in particular their effective rank --- influences the network learning regime. Through both empirical and theoretical analyses, we discover that high-rank initializations typically yield smaller network changes indicative of lazier learning, a finding we also confirm with experimentally-driven initial connectivity in recurrent neural networks. Conversely, low-rank initialization biases learning towards richer learning. Importantly, however, as an exception to this rule, we find lazier learning can still occur with a low-rank initialization that aligns with task and data statistics. Our research highlights the pivotal role of initial weight structures in shaping learning regimes, with implications for metabolic costs of plasticity and risks of catastrophic forgetting.  

\end{abstract}

\section{Introduction}

Structural variations can significantly impact learning dynamics in theoretical neuroscience studies of animals. For instance, studies have revealed that specific neural connectivity patterns can facilitate faster learning of certain tasks \citep{braun2022exact,raman2021frozen,simard2005fastest,canatar2021spectral,xie2022task,goudar2023schema,chang2023novo}. In deep learning, structure, encompassing architecture and initial connectivity, crucially dictates learning speed and effectiveness \citep{richards2019deep,zador2019critique,yang2021towards,braun2022exact}. 

A key structural aspect is the initial connectivity prior to training. Specifically, the initial connection weight magnitude can significantly bias learning dynamics, pushing them towards either rich or lazy regimes \citep{chizat2019lazy,flesch2021rich}. Lazy learning often induces minor changes in the network during the learning process. Such minimal adjustments are advantageous given that plasticity is metabolically costly \citep{mery2005cost, placcais2013favor}, and significant changes in representations might lead to issues like catastrophic forgetting \citep{mccloskey1989catastrophic, kirkpatrick2017overcoming}. 
On the other hand, the rich learning regime can significantly adapt the network's internal representations to task statistics, which can be advantageous for task feature acquisition and has implications for generalization \citep{flesch2021rich,george2022lazy}.  
Most research on initial weight magnitude's role in learning dynamics has focused on random Gaussian or Uniform initializations \citep{woodworth2020kernel,flesch2021rich,braun2022exact}. These patterns stand in contrast to the connectivity structures observed in biological neural circuits, which could exhibit a more pronounced low-rank eigenstructure \citep{song2005highly}. This divergence prompts a pivotal question: how does the initial weight structure, given a fixed initial weight magnitude, bias the learning regime?

This study examines how initial weight structure, particularly the effective rank, modulates the \textit{effective} richness or laziness of task learning within the standard training regime. We note that {\it rich} and {\it lazy} learning regimes have well established meanings in deep learning theory. The latter being defined as a situation where the Neural Tangent Kernel (NTK) stays stationary during training, while the former refers to the case where the NTK changes. In this work, we slightly extend these definitions and introduce \textbf{effective learning richness/laziness}. Unlike the traditional definition, which is based upon initialization, effective learning richness/laziness is defined in terms of post-training adjustment measurements. From this perspective, a learning process is deemed effectively "lazy" if the measured NTK movement is small. For example, consider a network whose initialization puts it in standard rich regime, but for a given task, its NTK moves very little during training. We define learning for this specific situation as effectively lazy. In other words, while the standard regime definition informs us (prior to training) whether the network can adapt significantly to task training or not, our "effective" definition lies in the post-training effects.  

\subsection{Contributions}

Our main \textbf{contributions} and findings can be summarized as follows:
\begin{itemize}
    \item Through theoretical derivation in two-layer feedforward linear network, we demonstrate that higher-rank initialization results in \textit{effectively} lazier learning \textbf{on average} across tasks (Theorem~\ref{thm:NTKalign}). We note that the emphasis of the theorem is on the expectation across tasks. 
    \item We validate our theoretical findings in recurrent neural networks (RNNs) through numerical experiments on well-known neuroscience tasks (Figure~\ref{fig:svd}) and demonstrate the applicability to different initial connectivity structures extracted from neuroscience data (Figure~\ref{fig:bio_spectrum_laziness}).
    \item We identify scenarios where certain low-rank initial weights still result in \textit{effectively} lazier learning for specific tasks (Proposition~\ref{prop:aligned_ini} and Figure~\ref{fig:aligned}). We postulate that such patterns emerge when a neural circuit is predisposed --- perhaps due to evolutionary factors or post-development --- to certain tasks, ingraining specific inductive biases in neural circuits.
\end{itemize}

\subsection{Related works}

An extended discussion on related works can also be found in Appendix~\ref{scn:related_works}.

\textbf{Theoretical Foundations of Neural Network Regimes and Implications for Neural Circuits:} The deep learning community has made tremendous strides in developing theoretical groundings for artificial neural networks \citep{advani2020high,jacot2018neural,alemohammad2020recurrent,agarwala2022second,atanasov2021neural,azulay2021implicit,emami2021implicit}. A focal point is the 'rich' and 'lazy' learning regimes dichotomy, which have distinct impacts on representation and generalization \citep{chizat2019lazy,flesch2021rich,geiger2020disentangling,george2022lazy,ghorbani2020neural,woodworth2020kernel,paccolat2021geometric,nacson2022implicit,haochen2021shape,flesch2023continual}. The 'lazy' regime results in minimal weight changes, while the 'rich' regime fosters task-specific adaptations. The transition between these is influenced by various factors, including initial weight scale and network width \citep{chizat2019lazy,geiger2020disentangling}.

Deep learning theories increasingly inform studies of biological neural network learning dynamics \citep{bordelon2022influence,liu2022beyond,braun2022exact,ghosh2023gradient,saxe2019information,farrell2022gradient,papyan2020prevalence,tishby2015deep}. For the rich/lazy regime theory, the existence of diverse learning regimes in neural systems is evident through the resource-intensive plasticity-driven transformations prevalent in developmental phases, followed by more subdued adjustments \citep{lohmann2014developmental}, and previous investigations characterized neural network behaviors under distinct regimes \citep{bordelon2022influence,schuessler2023aligned} and discerning which mode yields solutions mimicking neural data \citep{flesch2021rich}. 
Our work extends these studies by examining how initial weight structures affect learning.

\textbf{Neural circuit initialization, connectivity patterns and learning:} Extensive research has explored the influence of various random initializations on deep network learning \citep{saxe2013exact,bahri2020statistical,glorot2010understanding,he2015delving,arora2019theory}. The literature predominantly focuses on random initialization, but actual neural structures exhibit markedly different connectivity patterns, such as Dale's law and enriched cell-type-specific connectivity motifs \citep{rajan2006eigenvalue,ipsen2020consequences,harris2022eigenvalue,dahmen2020strong,aljadeff2015transition}. Motivated by existing evidence of low-rankedness in the brain \citep{thibeault2024low} and the overrepresentation of local motifs in neural circuits \citep{song2005highly}, which could be indicative of low-rank structures due to their influence on the eigenspectrum \citep{dahmen2020strong, shao2023relating}, our study explores the impact of connectivity effective rank on learning regimes. This focus is driven by the plausible presence of such low-rank structures in the brain, potentially revealed through these local motifs. With emerging connectivity data \citep{campagnola2022local,microns2021functional,dorkenwald2022flywire,winnubst2019reconstruction,scheffer2020connectome}, future work is poised to encompass rich additional features of connectivity.

\section{Setup and Theoretical Findings}

\subsection{RNN setup}

We examine recurrent neural networks (RNNs) because they are commonly adopted for modeling neural circuits \citep{barak2017recurrent, song2016training}. We consider a RNN with $N_{in}$ input units, $N$ hidden units and $N_{out}$ readout units (Figure~\ref{fig:svd}A). The update formula for $h_t \in \mathbb{R}^N$ (the hidden state at time $t$) is governed by \citep{ehrlich2021psychrnn,molano2022neurogym}:
\begin{equation}
    h_{t+1}= \rho h_t + (1-\rho) (W_h f(h_t) +  W_x x_t), \label{eqn:vrnn}
\end{equation}
where an exponential Euler approximation is made with $\rho = e^{-dt/\tau_m} \in \mathbb{R}$ denoting the leak factor for simulation time step $dt$ and $\tau_m$ denoting the membrane time constant; $f(\cdot): \mathbb{R}^N \rightarrow \mathbb{R}^N$ is the activation function, for which we use $ReLU$; $W_h \in \mathbb{R}^{N \times N}$ (resp. $W_x \in \mathbb{R}^{N \times N_{in}}$) is the recurrent (resp. input) weight matrix and $x_t \in \mathbb{R}^{N_{in}}$ is the input at time step $t$. Readout $\hat y_t \in \mathbb{R}^{N_{out}}$, with readout weights $w \in \mathbb{R}^{N_{out} \times N}$, is defined as
\begin{equation}
    \hat y_t = \langle w, f(h_t)\rangle. 
\end{equation}

The objective is to minimize scalar loss $L \in \mathbb{R}$, for which we use the cross-entropy loss for classification tasks and mean squared error for regression tasks. $L$ is minimized by updating the parameters using variants of gradient descent:
\begin{align}
    \Delta W &= -\eta \nabla_{W} L,
\end{align}
for learning rate $\eta \in \mathbb{R}$ and $W = [W_h \quad W_x \quad w^T] \in \mathbb{R}^{N \times (N_{in}+N+N_{out})}$ contains all the trainable parameters. Details of parameter settings can be found in Appendix~\ref{scn:sim_details}. 

\subsection{Effective laziness measures} \label{scn:laziness_measures}

As mentioned above, we introduce \textit{effective} richness and laziness, with effectively lazier (resp. richer) learning corresponding to less (resp. greater) network change over the course of learning. To quantify network change, we adopt the following three measures that have been used previously \citep{george2022lazy}. We note that these measures can be sensitive to other architectural aspects that bias learning regimes, such as network width, so throughout we hold these variables constant when making the comparisons.  

\textbf{Weight change norm} quantifies the vector norm of change in $W$. Effectively lazier learning should result in a lower weight change norm, and it is quantified as:
\begin{equation}
    \| \Delta W \| :=  \| W^{(f)} - W^{(0)} \|, 
\end{equation}
where $\| \cdot \| = \| \cdot \|_F$; $W^{(0)}$ (resp. $W^{(f)}$) are the weights before (resp. after) training. 
    
\textbf{Representation alignment (RA)} quantifies the directional change in a representational similarity matrix (RSM) before and after training. RSM focuses on the similarity between how two pairs of input are represented by computing the Gram matrix $R$ of last step hidden activity. Greater representation alignment indicates effectively lazier learning in the network, and it is obtained by
\begin{align}
    RA(R^{(f)}, R^{(0)}) &:= \frac{Tr(R^{(f)} R^{(0)})}{\|R^{(f)} \| \| R^{(0)}\|}, \quad \text{where } R := H^T H,
\end{align}
where $H \in \mathbb{R}^{N \times m}$ is the hidden activity at the last time step; $R^{(0)}$ and $R^{(f)} \in \mathbb{R}^{m \times m}$ are the initial and final RSM, respectively; $m$ is the batch size. 

\textbf{Tangent kernel alignment (KA)} quantifies the directional change in the neural tangent kernel (NTK) before and after training; effectively lazier learning should result in higher tangent kernel alignment.  The NTK computes the Gram matrix $K$ of the output gradient. Greater tangent kernel alignment points to effectively lazier learning, and it is obtained by
\begin{align} \label{eqn:NTKalign}
KA(K^{(f)}, K^{(0)}) &:= \frac{Tr(K^{(f)} K^{(0)})}{\|K^{(f)} \| \| K^{(0)}\|}, \quad \text{where } K := \nabla_W \hat y^T \nabla_W \hat y
\end{align}
where $K^{(0)}$ and $K^{(f)} \in \mathbb{R}^{m \times m}$ (for the $N_{out}=1$ case) denote the initial and final NTK, respectively.

\subsection{Theoretical findings} \label{scn:theory}

This subsection derives the theoretical impact of initial weight effective rank on tangent kernel alignment. First, Theorem~\ref{thm:NTKalign} focuses on \textbf{task-agnostic} settings, treating task definition as random variables and computing the \textbf{expected} tangent kernel alignment across tasks. With some assumptions, tangent kernel alignment is maximized when the initial weight singular values are distributed across all dimensions (i.e. high-rank initialization). 

In this section, our theoretical results are framed in a simplified feedforward setting, as we use a two-layer network with linear activations. However, we return to RNNs (Eq.~\ref{eqn:vrnn}) for the rest of the paper, and verify the generality of our theoretical findings with numerical experiments for both feedforward and recurrent architectures. Our choice is motivated by the need for theoretical tractability. While research on RNN learning in the NTK regime exists~\citep{yang2020tensor, alemohammad2020recurrent, emami2021implicit}, we are not aware of any studies featuring the final converged NTK that could serve as a basis for our comparison of the initial and final kernel. Consequently, we have chosen to focus on RNNs for neural circuit modeling and employ linear feedforward networks for theoretical derivations, a strategy also adopted by \citet{farrell2022gradient}; numerous other studies, including \citet{saxe2019information}, \citep{atanasov2021neural}, \citep{arora2019theory}, and \citep{braun2022exact}, have similarly concentrated on extracting theoretical insights from linear feedforward networks.  

 For a two-layer linear network with input data $X \in \mathbb{R}^{d \times m}$, $W_1 \in \mathbb{R}^{N \times d}$ and $W_2 \in \mathbb{R}^{1 \times N}$ as weights for layers 1 and 2, respectively, the NTK throughout training, $K$, is:
\begin{align}
    K = X^T (W_1^T W_1 + \| W_2 \|^2 I ) X. \label{eqn:NTK}
\end{align}
Without the loss of generality, suppose the output target $Y \in \mathbb{R}^{1 \times m}$ is generated from a linear teacher network as $Y = \beta^T X$, for some Gaussian vector
$\beta \in \R^d$, with $\beta_i \sim  \mathcal{N}(0, 1/d)$.

\begin{thrm} \label{thm:NTKalign}
    (Informal) Consider the network above with its corresponding NTK in Eq.~\ref{eqn:NTK}, trained under MSE loss with small initialization and whitened data.
    The expected kernel alignment across tasks is maximized with high-rank initialization, i.e. the singular values of $W^{(0)}_1$ are distributed across all dimensions. (Formal statement and proof are in Appendix~\ref{scn:proofs}) 
\end{thrm}


The intuition of Theorem~\ref{thm:NTKalign} result is that, when two random vectors are drawn in high-dimensional spaces, corresponding to the low-rank initial network and the task, the probability of them being nearly orthogonal is very high; this then necessitates greater movement to eventually learn the task direction. We emphasize again that Theorem~\ref{thm:NTKalign} is \textbf{task-agnostic}, i.e. it focuses on the \textbf{expected} tangent kernel alignment across input-output definitions.  This is in contrast to \textbf{task-specific} settings (e.g. \citet{woodworth2020kernel}) that focus on a given task.  In such task-specific settings, certain low-rank initializations can in fact lead to lazier learning. The following proposition predicts that if the task structure is known, low-rank initialization that is already aligned with the task statistics (input/output covariance) can lead to kernel alignment. We revisit this proposition again in Figure~\ref{fig:aligned}. We remark that initializing this way can still have high initial error because of randomized $W^{(0)}_2$.

\begin{prop} \label{prop:aligned_ini}
    (Informal) Following the setup and assumptions in Theorem~\ref{thm:NTKalign},  rank-1 
    initializations of the form $W^{(0)}_1 = \sigma [\beta^T/\|\beta\| \quad \vec{0} \quad ... \quad \vec{0}]$ leads to a high tangent kernel alignment. (Formal statement and proof are in Appendix~\ref{scn:proofs})
\end{prop}

Above, we state technical results in terms of one metric of the effective laziness of learning --- based on the NTK; our proof in Appendix~\ref{scn:proofs} easily extends also to the representation alignment metric.  
The impact on weight change is also assessed in Appendix Proposition~\ref{prop:change_requirement}. This is in line with our simulations with RNNs, which will show similar trends for all three of the metrics introduced in Section~\ref{scn:laziness_measures}).  

\section{Simulation results} \label{scn:sim_results}

\begin{figure}[h!]
    \centering
    \includegraphics[width=0.99\textwidth]{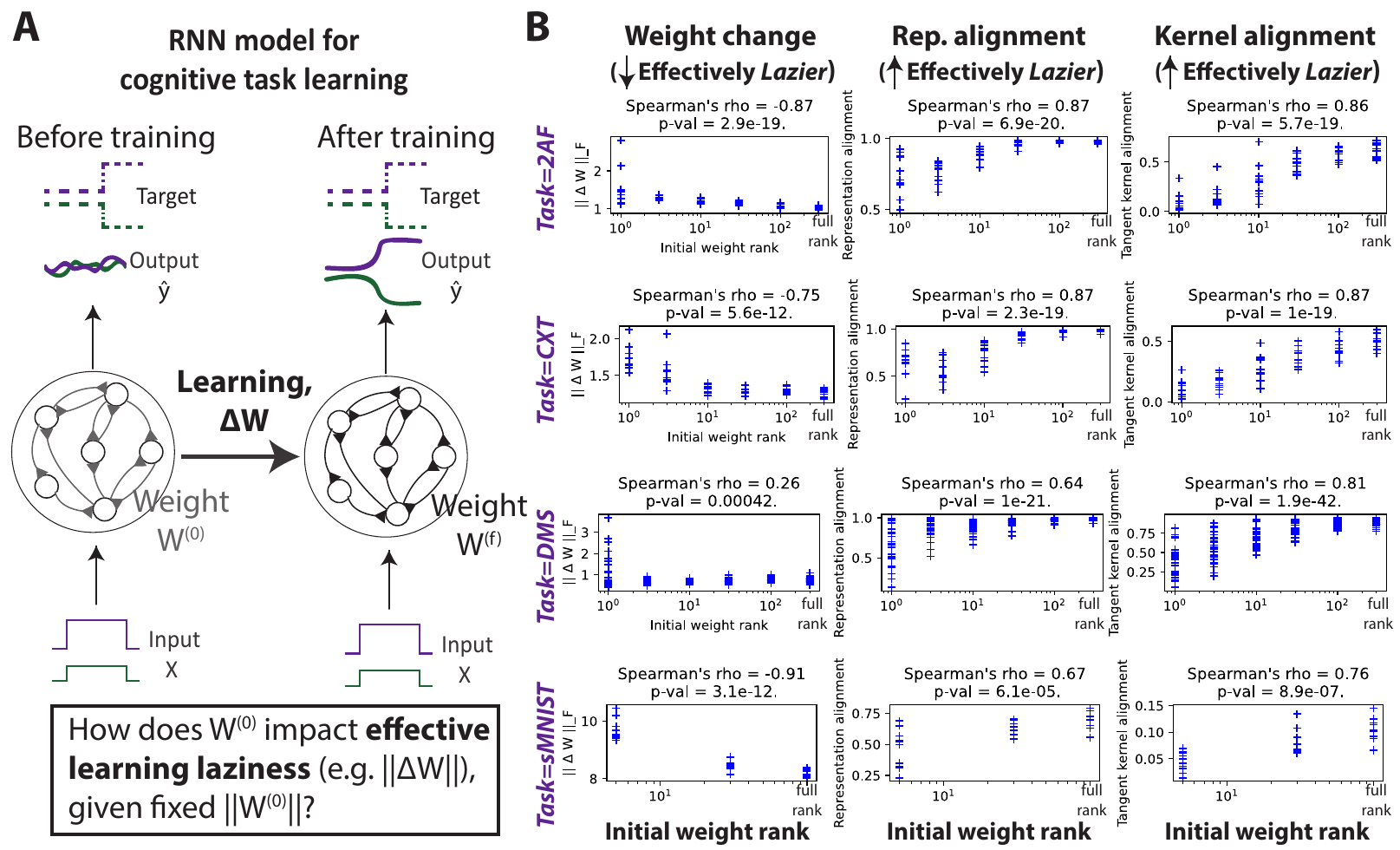}
    \caption{\textbf{Low-rank initial recurrent weights, generated using SVD, lead to greater changes (or effectively richer learning) in the recurrent neural network}. A) Schematic of RNN training setup. B)  Measurements of effective richness vs laziness of learning (metrics as defined in Section~\ref{scn:laziness_measures}), for RNN trained on several cognitive tasks in Neurogym \citep{molano2022neurogym} as well as the sequential MNIST task (sMNIST). For details on SVD weight creation, see Appendix~\ref{scn:sim_details}. Fewer rank points were used for sMNIST due to computational time. Each dot represents a single training run, with each run using a different random initialization (10 runs total for each setting). 
    } 
    \label{fig:svd}
\end{figure}

\begin{figure}[h!]
    \centering
    \includegraphics[width=0.99\textwidth]{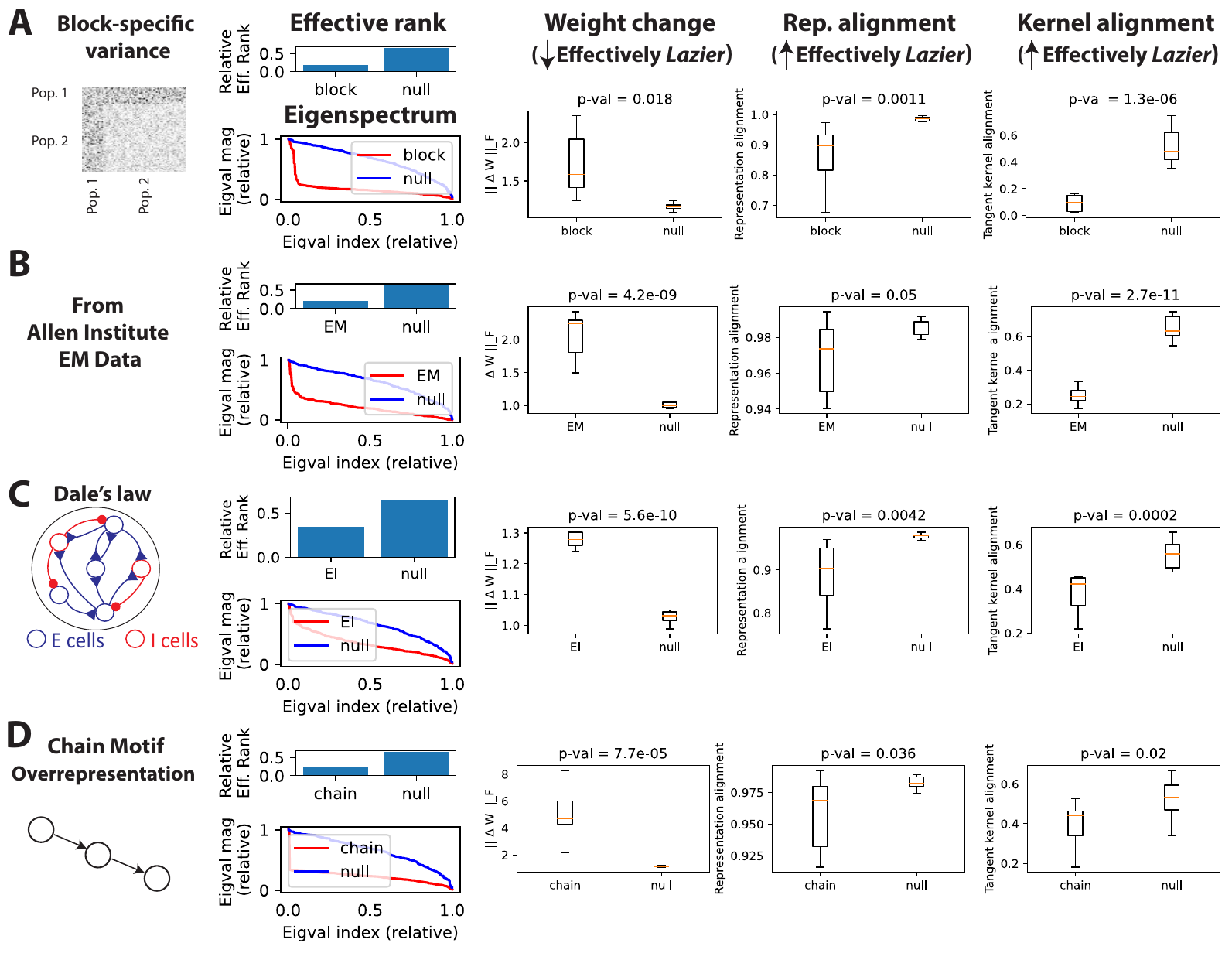}
    \caption{ \textbf{Low-rank initial weight structures, inspired by biological examples, lead to effectively richer learning}. We present the eigenspectrum and the relative effective rank of connectivity in A) structures with cell-type-specific statistics, B) structures derived from EM data, C) structures obeying Dale's law, and D) structures with an over-representation of chain motifs; we also present the effective learning laziness for networks initialized with these connectivity structures. These structures exhibit a lower effective rank compared to standard random Gaussian initialization (null). We plotted the magnitude of the eigenvalues (Eigval mag) --- scaled by the dominant eigenvalue's magnitude --- against their indices normalized by the network size $N$ (Eigval index). We apply the effective laziness measures described in Section~\ref{scn:laziness_measures} to compare the effective laziness of experimentally-driven initial connectivity versus standard random Gaussian initialization (null). See Appendix~\ref{scn:sim_details} for details on network initialization. The boxplots are generated from 10 independent runs with different initialization seeds. Due to space constraints, we include only the 2AF task here, but Appendix Figures~\ref{fig:bio_laziness_DMS} and~\ref{fig:bio_laziness_CXT} show that similar trends hold for the DMS and CXT tasks. 
    } 
    \label{fig:bio_spectrum_laziness}
\end{figure}

In this section we empirically illustrate and verify our main theoretical results, which are: \textbf{(1)} on average, high-rank initialization leads to effectively lazier learning (Theorem~\ref{thm:NTKalign});\textbf{ (2)} it is still possible for certain low-rank initializations that are already aligned to the task statistics to achieve effectively lazier learning (Proposition~\ref{prop:aligned_ini}).   

\begin{figure}[h!]
    \centering
    \includegraphics[width=0.99\textwidth]{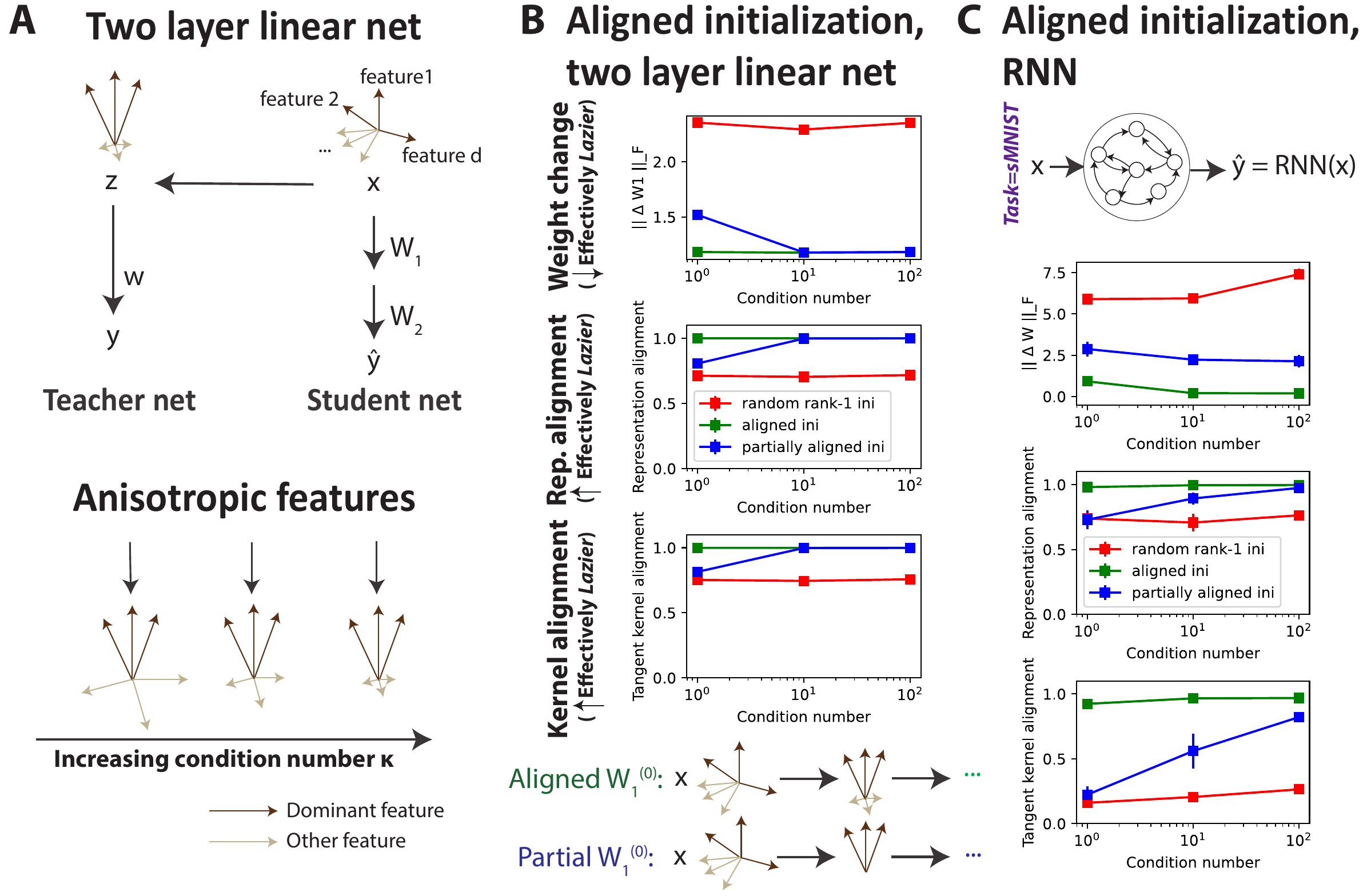}
    \caption{ \textbf{Low-rank initializations can still achieve high alignment for specific tasks (see Proposition~\ref{prop:aligned_ini}).}
A) The student-teacher two-layer linear network setup as described in Section~\ref{scn:theory}, but with feature anisotropy controlled by a feature modulation matrix $F$, i.e. $z=Fx$. The condition number of $F$ dictates the relative feature strength. We set the top half of the singular values of $F$ are set to $\kappa$, while the bottom half are set to $1$, where $\kappa$ represents the condition number of $F$.  B) The aligned initialization (green) is achieved by setting $W_1$ as described in Proposition~\ref{prop:aligned_ini} (with $\beta = w^T F$, $w$ is as illustrated), so that the initialization aligns with the task statistics. The partial alignment (blue) mirrors the aligned case, but $F$ is substituted with its rank-$(d/2)$ truncation, causing the network to align only with the dominant features. \textbf{We observe that a considerably higher alignment can be achieved when the initialization aligns solely with the dominant features, especially when
the relative strength of these dominant features is high.} 
C) The analysis from B) is replicated for RNNs learning the sMNIST task. As the ground truth network function is elusive, we use a teacher network with pre-trained weights. Once again, we replace $F$ with its rank-$(d/2)$ truncation for partial alignment. Details on the input/output definitions and initializations, as well as other simulation specifics, are available in Appendix~\ref{scn:sim_details}. We note that in all scenarios presented here, the initial errors are high since the readout weights are initialized randomly, rendering it a valid learning problem. 
    } 
    \label{fig:aligned}
\end{figure}

\textbf{Impact on effective laziness by low-rank initialization via SVD in RNNs:} As a proof-of-concept, we start in Figure~\ref{fig:svd} with low-rank initialization in RNNs by truncating an initial Gaussian random matrix via Singular Value Decomposition (SVD), which enables us to precisely control the rank, and rescale it to ensure that the comparison is across the same weight magnitude \citep{schuessler2020interplay}. Additionally, all comparisons were made after training was completed, and all these training sessions achieved comparable losses. For our investigations, we applied this initialization scheme across a variety of cognitive tasks --- including two-alternative forced choice (2AF), delayed-match-to-sample (DMS), context-dependent decision-making (CXT) tasks --- implemented with Neurogym \citep{molano2022neurogym} and the well-known machine learning benchmark sequential MNIST (sMNIST). Figure~\ref{fig:svd} indicates that low-rank initial weights result in effectively richer learning and greater network changes. 

These numerical trends are in line with Theorem~\ref{thm:NTKalign}, which focused on an idealized setting of a two-layer linear network with numerical results in Appendix Figure~\ref{fig:linear2layer}A. We also demonstrated this trend for a non-idealized feedforward setting in Appendix Figure~\ref{fig:linear2layer}B, and more explorations in feedforward settings and across a broader range of architecture is left for future exploration due to our focus on RNNs. In the Appendix, we show the main trends observed in Figure~\ref{fig:svd} also hold for Uniform initialization (Figure~\ref{fig:svd_uniform}), soft initial weight rank (Figure~\ref{fig:soft_rank}), various network sizes (Figure~\ref{fig:svd_varN}), learning rates (Figure~\ref{fig:svd_varLR}), gains (Figure~\ref{fig:svd_varWsig}), and finer time step $dt$ (Figure~\ref{fig:tau25_vs_laziness}). We note that, in addition to fixing the weight magnitude across comparisons, the dynamical regime might also confound learning regimes. A common method for controlling the dynamical regime is through the leading weight eigenvalue, which affects the top Lyapunov exponent. Controlling in this manner led to similar trends (Appendix Figure~\ref{fig:DomEig_vs_laziness}). Investigating the relationship between learning regimes and various concepts of dynamical regimes further is a promising direction for future work. Moreover, since our emphasis is on the effective learning regime, which is based on post-training changes, we concentrated on the laziness measures computed from networks after training, rather than during the learning process. However, we also tracked the alignment with the initial kernel and task kernel alignment during training (Appendix Figure~\ref{fig:Alignment_over_training}). We also examined how the kernel's effective rank evolves throughout the training period (Appendix Figure~\ref{fig:EffRank_over_training}). 

\textbf{Low-rank initialization via biologically motivated connectivity in RNNs:} To establish a closer connection with biological neural circuits, we have tested our predictions on low-rank initialization using a variety of biologically motivated structures capable of resulting in low-rank connectivity. Here are some of the examples: (A) connectivity with cell-type-specific statistics \citep{aljadeff2015transition}, where each block in the weight matrix corresponds to the connections between neurons of two distinct cell types, with the variance of these connections differing from one block to another. In terms of block-specific connectivity statistics, there are infinite possibilities for defining the blocks, each resulting in a unique eigenspectrum. For the example provided here, we adopted the setup from Figure S3 in \citet{aljadeff2015transition}, with parameters set as $\alpha=0.02$, $\gamma=10$, and $1-\epsilon=0.8$; these correspond to the fraction of hyperexcitable neurons, gain of hyperexcitable connections and gain of the rest, respectively. We follow this particular setup because it has been demonstrated to create an outlier leading eigenvalue, thereby reducing the effective rank. We also consider (B) connectivity matrix derived from the electron microscopy (EM) data \citep{alleninstitute2023brain}, where the synaptic connections between individual neurons are meticulously mapped to create a detailed and comprehensive representation of neural circuits. Also, we consider (C) connectivity obeying Dale's law, where each neuron is either excitatory or inhibitory, meaning it can only send out one type of signal – either increasing or decreasing the activity of connected neurons – a principle inspired by the way neurons behave in biological systems \citep{song2005highly}. Additionally, (D) the over-representation of certain localized connectivity patterns (or network motifs) --- such as the chain motif, where two cells are connected via a third intermediary cell --- creates outliers in the weight eigenspectrum, subsequently lowering the effective rank \citep{zhao2011synchronization,hu2018feedback,dahmen2020strong}. Details of these initial connectivity structures are provided in Appendix~\ref{scn:sim_details}. 

As illustrated in Figure~\ref{fig:bio_spectrum_laziness}, these connectivity structures, motivated by known features of biological neural networks, exhibit a lower effective rank compared to standard random Gaussian initialization, thereby serving as natural testbeds for our theoretical predictions. To quantify (relative) effective rank, we used $(\sum_i |\lambda_i |) / (| \lambda_1 | N)$, which indicates the fraction of eigenvalues on the order of the dominant one and captures the (scaled) area under the curve of the eigenspectrum plots. We also tried effective rank based on singular values, i.e. $(\sum_i |s_i |) / (| s_1 | N)$, in Appendix Figure~\ref{fig:Bio_spectrum_sval} and observed similar trends. Importantly, Figure~\ref{fig:bio_spectrum_laziness} show that these different low-rank biologically motivated structures can lead to effectively richer learning compared to the standard random Gaussian initialization.  This finding supports our overarching prediction, that lower rank initial weights leads to effectively richer learning. We note that to test our theoretical predictions based on gradient-descent learning without specific constraints on the solutions, the structures are enforced only at initialization and not constrained during training. In Appendix Figure~\ref{fig:ConstrainEI_vs_laziness}, we also constrained Dale's Law throughout training and found similar trends.

\textbf{Low-rank initialization aligned with task statistics:}  These simulations may be considered to be within our task-agnostic framework.  That is, we have chosen a ``random'' battery of tasks that is not directly matched to the initial network connectivity structures.  Thus, our findings that lower rank initializations lead to richer learning are expected from our theoretical prediction on the task-averaged alignment (Theorem~\ref{thm:NTKalign}), rather than something task-specific. However, Proposition~\ref{prop:aligned_ini} also predicts that low-rank initialization can lead to lazy learning if the initialization is already aligned to the task structure.   
To test this, we observe in Figure~\ref{fig:aligned} that a considerably higher alignment can be achieved when the initialization aligns solely with the dominant task features, especially when the relative strength of these dominant features is high. We postulate that such alignment may occur in biological settings if the circuit has evolved to preferentially learn specific tasks. 

\section{Discussion}

Our investigation casts light on the nuanced influence of initial weight effective rank on learning dynamics. 
Anchored by Theorem~\ref{thm:NTKalign}, our theoretical findings underscore that high-rank random initialization generally facilitates \textit{effectively} lazier learning on average across tasks. This focus on the expectation across tasks can provide insights into the circuit's flexibility in learning across a broad range of tasks as well as predict the effective learning regime when the task structure is uncertain. However, certain low-rank initial weights, when naturally predisposed to specific tasks, may lead to effectively lazier learning, suggesting an interesting interplay between evolutionary or developmental biases and learning dynamics (Proposition~\ref{prop:aligned_ini}). Our numerical experiments on RNNs further validate these theoretical findings illustrating the impact of initial rank in diverse settings. 

\textbf{Potential implications to neuroscience:} We investigate the impact of effective weight rank on learning regimes due to its relevance in neuroscience. Learning regimes reflect the extent of change through learning, implicating metabolic costs and catastrophic forgetting \citep{mccloskey1989catastrophic, placcais2013favor, mery2005cost}. The presence of different learning regimes is demonstrated in neural systems, since during developmental phases where neural circuits undergo extensive, plasticity-driven transformations. In contrast, mature neural circuits exhibit more subtle synaptic adjustments \citep{lohmann2014developmental}. We hypothesize that a circuit's task-specific alignment might be established either evolutionarily or during early development. The specialization of neural circuits, such as ventral versus dorsal \citep{bakhtiari2021functional}, may arise from engaging in tasks with similar computational demands. Conversely, circuits with high-rank structures may be less specialized, handling a wider array of tasks. Our framework could be used to compare connectivities across brain regions and species in order to predict their function and flexibility, assessing their functional specialization based on effective connectivity rank. Additionally, our framework predicts that connectivity rank will affect the degree of change in neural activity during the learning of new tasks. This hypothesis could be tested through BCI experiments, as shown in \citet{sadtler2014neural} and \citet{golub2018learning}, to explore how learning dynamics vary with connectivity rank.

Regarding deep learning, low-rank initialization is not a common practice, yet adaptations like low-rank updates have gained popularity in training large models \citep{hu2021lora}. LoRA, the study cited, concentrates on parameter updates rather than initializations, but understanding how update rank affects learning regimes is crucial. Our results offer a starting point for further exploration in this area. Although different rank initializations are less explored, with some exceptions like \citet{vodrahalli2022nonlinear}, our findings suggest that this area should receive more attention due to its potential effects on learning regimes and, consequently, on generalization \citep{george2022lazy}.

\textbf{Limitations and future directions:} Our study predominantly focused on the weight (effective) rank, leaving the exploration of other facets of weight on the effective learning regime as an open avenue. Also, the ramifications of effective learning regimes on learning speed --- given the known results on kernel alignment and ease of learning \citep{bartlett2021deep} and present mixed findings in the existing literature \citep{flesch2021rich, george2022lazy} --- warrant further exploration. 
 
Expanding the scope of our study calls for examining a wider variety of tasks, neural network architectures, and learning rules. Although our work is based on the backpropagation learning rule, its implications for biologically plausible learning rules remain unexplored. Our primary criterion for selecting tasks was their relevance to neuroscience, aligning with our main objectives. However, given the diverse range of tasks performed by various species, future research could benefit from exploring a more extensive array of tasks. Exploring more complex neuroscience tasks, such as those in Mod-Cog \citep{khona2023winning}, could provide valuable insights. On that note, we tested the pattern generation task from \citet{bellec2020solution}, a neuroscience task differing in structure from the Neurogym tasks, and observed similar trends (refer to Appendix Figure~\ref{fig:PG_laziness}). 
 
 Additionally, we ensured the consistency of outcomes against factors like width, learning rate, and initial gain (see Appendix~\ref{scn:more_sims}), but other factors such as dynamical regime and noise \citep{haochen2021shape} remain underexamined. On that note, the study’s focus on RNNs with finite task duration prompts further investigation into the implications for tasks with extended time steps and how conclusions for feedforward network depth \citep{xiao2020disentangling, seleznova2022neural} translate to RNN sequence length. Examining several mechanisms at once is beyond the scope of one paper, but our theoretical work constitutes the foundation for future investigations.
 
 Moreover, it is crucial to further explore the neuroscientific implications of effective learning regimes, as well as their diverse impacts on aspects such as representation, including kernel-task alignment (see Appendix Figure~\ref{fig:Alignment_over_training}), and generalization capabilities \citep{flesch2021rich, george2022lazy, schuessler2023aligned}. Our current study did not delve into how initial weight rank affects these facets of learning, representing an essential future direction in connecting weight rank to these theoretical implications in both biological and artificial neural networks. 

Furthermore, while there exists evidence for low-rankedness in the brain \citep{thibeault2024low}, the extent to which the brain uses low-rank structures remains an open question, especially as neural circuit structures can vary across regions and species. While local connectivity statistics~\citep{song2005highly} can offer some predictive insight into the global low-rank structure, this relationship is not always immediately apparent~\citep{shao2023relating}. Our theoretical results contribute to understanding the role of connectivity rank in the brain by linking effective connectivity rank with learning dynamics. 
 
 Lastly, we have primarily examined low-rank tasks and there remains unexplored terrain regarding the interplay between the number of task classes and weight rank, which is pivotal to uncovering a more precise relationship between the effective learning regime and the initial weight rank \citep{dubreuil2022role, gao2017theory}. Overall, this dynamic area of learning regimes is ripe for many explorations, integrating numerous factors; our work contributes to this exciting area with new tools.

\section{Acknowledgement}
We thank Andrew Saxe, Stefano Recanatesi, Kyle Aitken and Dana Mastrovito for insightful discussions and helpful feedback. This research was supported by NSERC PGS-D (Y.H.L.); FRQNT B2X (Y.H.L.); Pearson Fellowship (Y.H.L.); NSF AccelNet IN-BIC program, Grant No. OISE-2019976 AM02 (Y.H.L.); NIH BRAIN, Grant No. R01 1RF1DA055669 (Y.H.L., E.S.B., S.M.); Mitacs Globalink Research Award (Y.H.L.);  IVADO Postdoctoral Fellowship (J.C); the Canada First Research Excellence Fund (J.C.); NSERC Discovery Grant RGPIN-2018-04821 (G.L); Canada Research Chair in Neural Computations and Interfacing (G.L.); Canada CIFAR AI Chair program (G.L.). We also thank the Allen Institute founder, Paul G. Allen, for his vision, encouragement, and support. 

\bibliography{ref_main}

\begin{thebibliography}{102}
\providecommand{\natexlab}[1]{#1}
\providecommand{\url}[1]{\texttt{#1}}
\expandafter\ifx\csname urlstyle\endcsname\relax
  \providecommand{\doi}[1]{doi: #1}\else
  \providecommand{\doi}{doi: \begingroup \urlstyle{rm}\Url}\fi

\bibitem[Advani et~al.(2020)Advani, Saxe, and Sompolinsky]{advani2020high}
Madhu~S Advani, Andrew~M Saxe, and Haim Sompolinsky.
\newblock High-dimensional dynamics of generalization error in neural networks.
\newblock \emph{Neural Networks}, 132:\penalty0 428--446, 2020.

\bibitem[Agarwala et~al.(2022)Agarwala, Pedregosa, and
  Pennington]{agarwala2022second}
Atish Agarwala, Fabian Pedregosa, and Jeffrey Pennington.
\newblock Second-order regression models exhibit progressive sharpening to the
  edge of stability.
\newblock \emph{arXiv preprint arXiv:2210.04860}, 2022.

\bibitem[Alemohammad et~al.(2020)Alemohammad, Wang, Balestriero, and
  Baraniuk]{alemohammad2020recurrent}
Sina Alemohammad, Zichao Wang, Randall Balestriero, and Richard Baraniuk.
\newblock The recurrent neural tangent kernel.
\newblock \emph{arXiv preprint arXiv:2006.10246}, 2020.

\bibitem[Aljadeff et~al.(2015)Aljadeff, Stern, and
  Sharpee]{aljadeff2015transition}
Johnatan Aljadeff, Merav Stern, and Tatyana Sharpee.
\newblock Transition to chaos in random networks with cell-type-specific
  connectivity.
\newblock \emph{Physical review letters}, 114\penalty0 (8):\penalty0 088101,
  2015.

\bibitem[{Allen Institute}(2023)]{alleninstitute2023brain}
{Allen Institute}.
\newblock Allen institute for brain science, 2023.
\newblock
  \url{https://portal.brain-map.org/explore/connectivity/ultrastructural-connectomics/}.

\bibitem[Arora et~al.(2019)Arora, Arora, Bruna, Cohen, Du, Ge, Gunasekar, Jin,
  Lee, Ma, et~al.]{arora2019theory}
Raman Arora, Sanjeev Arora, Joan Bruna, Nadav Cohen, Simon Du, Rong Ge, Suriya
  Gunasekar, Chi Jin, Jason Lee, Tengyu Ma, et~al.
\newblock Theory of deep learning, 2019.

\bibitem[Atanasov et~al.(2021)Atanasov, Bordelon, and
  Pehlevan]{atanasov2021neural}
Alexander Atanasov, Blake Bordelon, and Cengiz Pehlevan.
\newblock Neural networks as kernel learners: The silent alignment effect.
\newblock \emph{arXiv preprint arXiv:2111.00034}, 2021.

\bibitem[Azulay et~al.(2021)Azulay, Moroshko, Nacson, Woodworth, Srebro,
  Globerson, and Soudry]{azulay2021implicit}
Shahar Azulay, Edward Moroshko, Mor~Shpigel Nacson, Blake~E Woodworth, Nathan
  Srebro, Amir Globerson, and Daniel Soudry.
\newblock On the implicit bias of initialization shape: Beyond infinitesimal
  mirror descent.
\newblock In \emph{International Conference on Machine Learning}, pp.\
  468--477. PMLR, 2021.

\bibitem[Bahri et~al.(2020)Bahri, Kadmon, Pennington, Schoenholz,
  Sohl-Dickstein, and Ganguli]{bahri2020statistical}
Yasaman Bahri, Jonathan Kadmon, Jeffrey Pennington, Sam~S Schoenholz, Jascha
  Sohl-Dickstein, and Surya Ganguli.
\newblock Statistical mechanics of deep learning.
\newblock \emph{Annual Review of Condensed Matter Physics}, 11:\penalty0
  501--528, 2020.

\bibitem[Bakhtiari et~al.(2021)Bakhtiari, Mineault, Lillicrap, Pack, and
  Richards]{bakhtiari2021functional}
Shahab Bakhtiari, Patrick Mineault, Timothy Lillicrap, Christopher Pack, and
  Blake Richards.
\newblock The functional specialization of visual cortex emerges from training
  parallel pathways with self-supervised predictive learning.
\newblock \emph{Advances in Neural Information Processing Systems},
  34:\penalty0 25164--25178, 2021.

\bibitem[Barak(2017)]{barak2017recurrent}
Omri Barak.
\newblock Recurrent neural networks as versatile tools of neuroscience
  research.
\newblock \emph{Current opinion in neurobiology}, 46:\penalty0 1--6, 2017.

\bibitem[Baratin et~al.(2021)Baratin, George, Laurent, Hjelm, Lajoie, Vincent,
  and Lacoste-Julien]{baratin2021implicit}
Aristide Baratin, Thomas George, C{\'e}sar Laurent, R~Devon Hjelm, Guillaume
  Lajoie, Pascal Vincent, and Simon Lacoste-Julien.
\newblock Implicit regularization via neural feature alignment.
\newblock In \emph{International Conference on Artificial Intelligence and
  Statistics}, pp.\  2269--2277. PMLR, 2021.

\bibitem[Bartlett et~al.(2021)Bartlett, Montanari, and
  Rakhlin]{bartlett2021deep}
Peter~L Bartlett, Andrea Montanari, and Alexander Rakhlin.
\newblock Deep learning: a statistical viewpoint.
\newblock \emph{Acta numerica}, 30:\penalty0 87--201, 2021.

\bibitem[Bellec et~al.(2020)Bellec, Scherr, Subramoney, Hajek, Salaj,
  Legenstein, and Maass]{bellec2020solution}
Guillaume Bellec, Franz Scherr, Anand Subramoney, Elias Hajek, Darjan Salaj,
  Robert Legenstein, and Wolfgang Maass.
\newblock A solution to the learning dilemma for recurrent networks of spiking
  neurons.
\newblock \emph{Nature communications}, 11\penalty0 (1):\penalty0 3625, 2020.

\bibitem[Bordelon \& Pehlevan(2022)Bordelon and
  Pehlevan]{bordelon2022influence}
Blake Bordelon and Cengiz Pehlevan.
\newblock The influence of learning rule on representation dynamics in wide
  neural networks.
\newblock \emph{arXiv preprint arXiv:2210.02157}, 2022.

\bibitem[Braun et~al.(2022)Braun, Domin{\'e}, Fitzgerald, and
  Saxe]{braun2022exact}
Lukas Braun, Cl{\'e}mentine Domin{\'e}, James Fitzgerald, and Andrew Saxe.
\newblock Exact learning dynamics of deep linear networks with prior knowledge.
\newblock \emph{Advances in Neural Information Processing Systems},
  35:\penalty0 6615--6629, 2022.

\bibitem[Campagnola et~al.(2022)Campagnola, Seeman, Chartrand, Kim, Hoggarth,
  Gamlin, Ito, Trinh, Davoudian, Radaelli, et~al.]{campagnola2022local}
Luke Campagnola, Stephanie~C Seeman, Thomas Chartrand, Lisa Kim, Alex Hoggarth,
  Clare Gamlin, Shinya Ito, Jessica Trinh, Pasha Davoudian, Cristina Radaelli,
  et~al.
\newblock Local connectivity and synaptic dynamics in mouse and human
  neocortex.
\newblock \emph{Science}, 375\penalty0 (6585):\penalty0 eabj5861, 2022.

\bibitem[Canatar et~al.(2021)Canatar, Bordelon, and
  Pehlevan]{canatar2021spectral}
Abdulkadir Canatar, Blake Bordelon, and Cengiz Pehlevan.
\newblock Spectral bias and task-model alignment explain generalization in
  kernel regression and infinitely wide neural networks.
\newblock \emph{Nature communications}, 12\penalty0 (1):\penalty0 2914, 2021.

\bibitem[Chang et~al.(2023)Chang, Perich, Miller, Gallego, and
  Clopath]{chang2023novo}
Joanna~C Chang, Matthew~G Perich, Lee~E Miller, Juan~A Gallego, and Claudia
  Clopath.
\newblock De novo motor learning creates structure in neural activity space
  that shapes adaptation.
\newblock \emph{bioRxiv}, pp.\  2023--05, 2023.

\bibitem[Chizat et~al.(2019)Chizat, Oyallon, and Bach]{chizat2019lazy}
Lenaic Chizat, Edouard Oyallon, and Francis Bach.
\newblock On lazy training in differentiable programming.
\newblock \emph{Advances in neural information processing systems}, 32, 2019.

\bibitem[Dahmen et~al.(2020)Dahmen, Recanatesi, Ocker, Jia, Helias, and
  Shea-Brown]{dahmen2020strong}
David Dahmen, Stefano Recanatesi, Gabriel~K Ocker, Xiaoxuan Jia, Moritz Helias,
  and Eric Shea-Brown.
\newblock Strong coupling and local control of dimensionality across brain
  areas.
\newblock \emph{Biorxiv}, pp.\  2020--11, 2020.

\bibitem[Diederich \& Opper(1987)Diederich and Opper]{diederich1987learning}
Sigurd Diederich and Manfred Opper.
\newblock Learning of correlated patterns in spin-glass networks by local
  learning rules.
\newblock \emph{Physical review letters}, 58\penalty0 (9):\penalty0 949, 1987.

\bibitem[Dorkenwald et~al.(2022)Dorkenwald, McKellar, Macrina, Kemnitz, Lee,
  Lu, Wu, Popovych, Mitchell, Nehoran, et~al.]{dorkenwald2022flywire}
Sven Dorkenwald, Claire~E McKellar, Thomas Macrina, Nico Kemnitz, Kisuk Lee,
  Ran Lu, Jingpeng Wu, Sergiy Popovych, Eric Mitchell, Barak Nehoran, et~al.
\newblock Flywire: online community for whole-brain connectomics.
\newblock \emph{Nature methods}, 19\penalty0 (1):\penalty0 119--128, 2022.

\bibitem[Dubreuil et~al.(2022)Dubreuil, Valente, Beiran, Mastrogiuseppe, and
  Ostojic]{dubreuil2022role}
Alexis Dubreuil, Adrian Valente, Manuel Beiran, Francesca Mastrogiuseppe, and
  Srdjan Ostojic.
\newblock The role of population structure in computations through neural
  dynamics.
\newblock \emph{Nature neuroscience}, 25\penalty0 (6):\penalty0 783--794, 2022.

\bibitem[Ehrlich et~al.(2021)Ehrlich, Stone, Brandfonbrener, Atanasov, and
  Murray]{ehrlich2021psychrnn}
Daniel~B Ehrlich, Jasmine~T Stone, David Brandfonbrener, Alexander Atanasov,
  and John~D Murray.
\newblock Psychrnn: An accessible and flexible python package for training
  recurrent neural network models on cognitive tasks.
\newblock \emph{Eneuro}, 8\penalty0 (1), 2021.

\bibitem[Emami et~al.(2021)Emami, Sahraee-Ardakan, Pandit, Rangan, and
  Fletcher]{emami2021implicit}
Melikasadat Emami, Mojtaba Sahraee-Ardakan, Parthe Pandit, Sundeep Rangan, and
  Alyson~K Fletcher.
\newblock Implicit bias of linear rnns.
\newblock In \emph{International Conference on Machine Learning}, pp.\
  2982--2992. PMLR, 2021.

\bibitem[Farrell et~al.(2022)Farrell, Recanatesi, Moore, Lajoie, and
  Shea-Brown]{farrell2022gradient}
Matthew Farrell, Stefano Recanatesi, Timothy Moore, Guillaume Lajoie, and Eric
  Shea-Brown.
\newblock Gradient-based learning drives robust representations in recurrent
  neural networks by balancing compression and expansion.
\newblock \emph{Nature Machine Intelligence}, 4\penalty0 (6):\penalty0
  564--573, 2022.

\bibitem[Flesch et~al.(2021)Flesch, Juechems, Dumbalska, Saxe, and
  Summerfield]{flesch2021rich}
Timo Flesch, Keno Juechems, Tsvetomira Dumbalska, Andrew Saxe, and Christopher
  Summerfield.
\newblock Rich and lazy learning of task representations in brains and neural
  networks.
\newblock \emph{BioRxiv}, pp.\  2021--04, 2021.

\bibitem[Flesch et~al.(2023)Flesch, Saxe, and Summerfield]{flesch2023continual}
Timo Flesch, Andrew Saxe, and Christopher Summerfield.
\newblock Continual task learning in natural and artificial agents.
\newblock \emph{Trends in Neurosciences}, 46\penalty0 (3):\penalty0 199--210,
  2023.

\bibitem[Gao et~al.(2017)Gao, Trautmann, Yu, Santhanam, Ryu, Shenoy, and
  Ganguli]{gao2017theory}
Peiran Gao, Eric Trautmann, Byron Yu, Gopal Santhanam, Stephen Ryu, Krishna
  Shenoy, and Surya Ganguli.
\newblock A theory of multineuronal dimensionality, dynamics and measurement.
\newblock \emph{BioRxiv}, pp.\  214262, 2017.

\bibitem[Geiger et~al.(2020)Geiger, Spigler, Jacot, and
  Wyart]{geiger2020disentangling}
Mario Geiger, Stefano Spigler, Arthur Jacot, and Matthieu Wyart.
\newblock Disentangling feature and lazy training in deep neural networks.
\newblock \emph{Journal of Statistical Mechanics: Theory and Experiment},
  2020\penalty0 (11):\penalty0 113301, 2020.

\bibitem[George et~al.(2022)George, Lajoie, and Baratin]{george2022lazy}
Thomas George, Guillaume Lajoie, and Aristide Baratin.
\newblock Lazy vs hasty: linearization in deep networks impacts learning
  schedule based on example difficulty.
\newblock \emph{TMLR}, 2022.

\bibitem[Ghorbani et~al.(2020)Ghorbani, Mei, Misiakiewicz, and
  Montanari]{ghorbani2020neural}
Behrooz Ghorbani, Song Mei, Theodor Misiakiewicz, and Andrea Montanari.
\newblock When do neural networks outperform kernel methods?
\newblock \emph{Advances in Neural Information Processing Systems},
  33:\penalty0 14820--14830, 2020.

\bibitem[Ghosh et~al.(2023)Ghosh, Liu, Lajoie, Kording, and
  Richards]{ghosh2023gradient}
Arna Ghosh, Yuhan~Helena Liu, Guillaume Lajoie, Konrad Kording, and Blake~Aaron
  Richards.
\newblock How gradient estimator variance and bias impact learning in neural
  networks.
\newblock In \emph{The Eleventh International Conference on Learning
  Representations}, 2023.

\bibitem[Glorot \& Bengio(2010)Glorot and Bengio]{glorot2010understanding}
Xavier Glorot and Yoshua Bengio.
\newblock Understanding the difficulty of training deep feedforward neural
  networks.
\newblock In \emph{Proceedings of the thirteenth international conference on
  artificial intelligence and statistics}, pp.\  249--256. JMLR Workshop and
  Conference Proceedings, 2010.

\bibitem[Golub et~al.(2018)Golub, Sadtler, Oby, Quick, Ryu, Tyler-Kabara,
  Batista, Chase, and Yu]{golub2018learning}
Matthew~D Golub, Patrick~T Sadtler, Emily~R Oby, Kristin~M Quick, Stephen~I
  Ryu, Elizabeth~C Tyler-Kabara, Aaron~P Batista, Steven~M Chase, and Byron~M
  Yu.
\newblock Learning by neural reassociation.
\newblock \emph{Nature neuroscience}, 21\penalty0 (4):\penalty0 607--616, 2018.

\bibitem[Goudar et~al.(2023)Goudar, Peysakhovich, Freedman, Buffalo, and
  Wang]{goudar2023schema}
Vishwa Goudar, Barbara Peysakhovich, David~J Freedman, Elizabeth~A Buffalo, and
  Xiao-Jing Wang.
\newblock Schema formation in a neural population subspace underlies
  learning-to-learn in flexible sensorimotor problem-solving.
\newblock \emph{Nature Neuroscience}, 26\penalty0 (5):\penalty0 879--890, 2023.

\bibitem[Greedy et~al.(2022)Greedy, Zhu, Pemberton, Mellor, and
  Ponte~Costa]{greedy2022single}
Will Greedy, Heng~Wei Zhu, Joseph Pemberton, Jack Mellor, and Rui Ponte~Costa.
\newblock Single-phase deep learning in cortico-cortical networks.
\newblock \emph{Advances in Neural Information Processing Systems},
  35:\penalty0 24213--24225, 2022.

\bibitem[HaoChen et~al.(2021)HaoChen, Wei, Lee, and Ma]{haochen2021shape}
Jeff~Z HaoChen, Colin Wei, Jason Lee, and Tengyu Ma.
\newblock Shape matters: Understanding the implicit bias of the noise
  covariance.
\newblock In \emph{Conference on Learning Theory}, pp.\  2315--2357. PMLR,
  2021.

\bibitem[Harris et~al.(2022)Harris, Meffin, Burkitt, and
  Peterson]{harris2022eigenvalue}
Isabelle~D Harris, Hamish Meffin, Anthony~N Burkitt, and Andre~DH Peterson.
\newblock Eigenvalue spectral properties of sparse random matrices obeying
  dale's law.
\newblock \emph{arXiv preprint arXiv:2212.01549}, 2022.

\bibitem[He et~al.(2015)He, Zhang, Ren, and Sun]{he2015delving}
Kaiming He, Xiangyu Zhang, Shaoqing Ren, and Jian Sun.
\newblock Delving deep into rectifiers: Surpassing human-level performance on
  imagenet classification.
\newblock In \emph{Proceedings of the IEEE international conference on computer
  vision}, pp.\  1026--1034, 2015.

\bibitem[Hinton(2022)]{hinton2022forward}
Geoffrey Hinton.
\newblock The forward-forward algorithm: Some preliminary investigations.
\newblock \emph{arXiv preprint arXiv:2212.13345}, 2022.

\bibitem[Hu et~al.(2021)Hu, Shen, Wallis, Allen-Zhu, Li, Wang, Wang, and
  Chen]{hu2021lora}
Edward~J Hu, Yelong Shen, Phillip Wallis, Zeyuan Allen-Zhu, Yuanzhi Li, Shean
  Wang, Lu~Wang, and Weizhu Chen.
\newblock Lora: Low-rank adaptation of large language models.
\newblock \emph{arXiv preprint arXiv:2106.09685}, 2021.

\bibitem[Hu et~al.(2018)Hu, Brunton, Cain, Mihalas, Kutz, and
  Shea-Brown]{hu2018feedback}
Yu~Hu, Steven~L Brunton, Nicholas Cain, Stefan Mihalas, J~Nathan Kutz, and Eric
  Shea-Brown.
\newblock Feedback through graph motifs relates structure and function in
  complex networks.
\newblock \emph{Physical Review E}, 98\penalty0 (6):\penalty0 062312, 2018.

\bibitem[Ipsen \& Peterson(2020)Ipsen and Peterson]{ipsen2020consequences}
Jesper~R Ipsen and Andre~DH Peterson.
\newblock Consequences of dale's law on the stability-complexity relationship
  of random neural networks.
\newblock \emph{Physical Review E}, 101\penalty0 (5):\penalty0 052412, 2020.

\bibitem[Jacot et~al.(2018)Jacot, Gabriel, and Hongler]{jacot2018neural}
Arthur Jacot, Franck Gabriel, and Cl{\'e}ment Hongler.
\newblock Neural tangent kernel: Convergence and generalization in neural
  networks.
\newblock \emph{Advances in neural information processing systems}, 31, 2018.

\bibitem[Khona et~al.(2023)Khona, Chandra, Ma, and Fiete]{khona2023winning}
Mikail Khona, Sarthak Chandra, Joy~J Ma, and Ila~R Fiete.
\newblock Winning the lottery with neural connectivity constraints: Faster
  learning across cognitive tasks with spatially constrained sparse rnns.
\newblock \emph{Neural Computation}, 35\penalty0 (11):\penalty0 1850--1869,
  2023.

\bibitem[Kirkpatrick et~al.(2017)Kirkpatrick, Pascanu, Rabinowitz, Veness,
  Desjardins, Rusu, Milan, Quan, Ramalho, Grabska-Barwinska,
  et~al.]{kirkpatrick2017overcoming}
James Kirkpatrick, Razvan Pascanu, Neil Rabinowitz, Joel Veness, Guillaume
  Desjardins, Andrei~A Rusu, Kieran Milan, John Quan, Tiago Ramalho, Agnieszka
  Grabska-Barwinska, et~al.
\newblock Overcoming catastrophic forgetting in neural networks.
\newblock \emph{Proceedings of the national academy of sciences}, 114\penalty0
  (13):\penalty0 3521--3526, 2017.

\bibitem[Laborieux \& Zenke(2022)Laborieux and Zenke]{laborieux2022holomorphic}
Axel Laborieux and Friedemann Zenke.
\newblock Holomorphic equilibrium propagation computes exact gradients through
  finite size oscillations.
\newblock \emph{arXiv preprint arXiv:2209.00530}, 2022.

\bibitem[LeCun(1998)]{lecun1998mnist}
Yann LeCun.
\newblock The mnist database of handwritten digits.
\newblock \emph{http://yann. lecun. com/exdb/mnist/}, 1998.

\bibitem[Lillicrap et~al.(2020)Lillicrap, Santoro, Marris, Akerman, and
  Hinton]{lillicrap2020backpropagation}
Timothy~P Lillicrap, Adam Santoro, Luke Marris, Colin~J Akerman, and Geoffrey
  Hinton.
\newblock Backpropagation and the brain.
\newblock \emph{Nature Reviews Neuroscience}, 21\penalty0 (6):\penalty0
  335--346, 2020.

\bibitem[Liu et~al.(2021)Liu, Smith, Mihalas, Shea-Brown, and
  S{\"u}mb{\"u}l]{liu2021cell}
Yuhan~Helena Liu, Stephen Smith, Stefan Mihalas, Eric Shea-Brown, and Uygar
  S{\"u}mb{\"u}l.
\newblock Cell-type--specific neuromodulation guides synaptic credit assignment
  in a spiking neural network.
\newblock \emph{Proceedings of the National Academy of Sciences}, 118\penalty0
  (51):\penalty0 e2111821118, 2021.

\bibitem[Liu et~al.(2022{\natexlab{a}})Liu, Ghosh, Richards, Shea-Brown, and
  Lajoie]{liu2022beyond}
Yuhan~Helena Liu, Arna Ghosh, Blake Richards, Eric Shea-Brown, and Guillaume
  Lajoie.
\newblock Beyond accuracy: generalization properties of bio-plausible temporal
  credit assignment rules.
\newblock \emph{Advances in Neural Information Processing Systems},
  35:\penalty0 23077--23097, 2022{\natexlab{a}}.

\bibitem[Liu et~al.(2022{\natexlab{b}})Liu, Smith, Mihalas, Shea-Brown, and
  S{\"u}mb{\"u}l]{liu2022biologically}
Yuhan~Helena Liu, Stephen Smith, Stefan Mihalas, Eric Shea-Brown, and Uygar
  S{\"u}mb{\"u}l.
\newblock Biologically-plausible backpropagation through arbitrary timespans
  via local neuromodulators.
\newblock \emph{arXiv preprint arXiv:2206.01338}, 2022{\natexlab{b}}.

\bibitem[Lohmann \& Kessels(2014)Lohmann and Kessels]{lohmann2014developmental}
Christian Lohmann and Helmut~W Kessels.
\newblock The developmental stages of synaptic plasticity.
\newblock \emph{The Journal of physiology}, 592\penalty0 (1):\penalty0 13--31,
  2014.

\bibitem[Marschall et~al.(2020)Marschall, Cho, and Savin]{marschall2020unified}
Owen Marschall, Kyunghyun Cho, and Cristina Savin.
\newblock A unified framework of online learning algorithms for training
  recurrent neural networks.
\newblock \emph{The Journal of Machine Learning Research}, 21\penalty0
  (1):\penalty0 5320--5353, 2020.

\bibitem[McCloskey \& Cohen(1989)McCloskey and
  Cohen]{mccloskey1989catastrophic}
Michael McCloskey and Neal~J Cohen.
\newblock Catastrophic interference in connectionist networks: The sequential
  learning problem.
\newblock In \emph{Psychology of learning and motivation}, volume~24, pp.\
  109--165. Elsevier, 1989.

\bibitem[Mery \& Kawecki(2005)Mery and Kawecki]{mery2005cost}
Frederic Mery and Tadeusz~J Kawecki.
\newblock A cost of long-term memory in drosophila.
\newblock \emph{Science}, 308\penalty0 (5725):\penalty0 1148--1148, 2005.

\bibitem[Meulemans et~al.(2022)Meulemans, Zucchet, Kobayashi, Von~Oswald, and
  Sacramento]{meulemans2022least}
Alexander Meulemans, Nicolas Zucchet, Seijin Kobayashi, Johannes Von~Oswald,
  and Jo{\~a}o Sacramento.
\newblock The least-control principle for local learning at equilibrium.
\newblock \emph{Advances in Neural Information Processing Systems},
  35:\penalty0 33603--33617, 2022.

\bibitem[{MICrONS Consortium} et~al.(2021){MICrONS Consortium}, Bae, Baptiste,
  Bishop, Bodor, Brittain, Buchanan, Bumbarger, Castro, Celii,
  et~al.]{microns2021functional}
{MICrONS Consortium}, J~Alexander Bae, Mahaly Baptiste, Caitlyn~A Bishop,
  Agnes~L Bodor, Derrick Brittain, JoAnn Buchanan, Daniel~J Bumbarger, Manuel~A
  Castro, Brendan Celii, et~al.
\newblock Functional connectomics spanning multiple areas of mouse visual
  cortex.
\newblock \emph{BioRxiv}, pp.\  2021--07, 2021.

\bibitem[Molano-Mazon et~al.(2022)Molano-Mazon, Barbosa, Pastor-Ciurana,
  Fradera, Zhang, Forest, del Pozo~Lerida, Ji-An, Cueva, de~la Rocha,
  et~al.]{molano2022neurogym}
Manuel Molano-Mazon, Joao Barbosa, Jordi Pastor-Ciurana, Marta Fradera, Ru-Yuan
  Zhang, Jeremy Forest, Jorge del Pozo~Lerida, Li~Ji-An, Christopher~J Cueva,
  Jaime de~la Rocha, et~al.
\newblock Neurogym: An open resource for developing and sharing neuroscience
  tasks.
\newblock 2022.

\bibitem[Murray(2019)]{murray2019local}
James~M Murray.
\newblock Local online learning in recurrent networks with random feedback.
\newblock \emph{Elife}, 8:\penalty0 e43299, 2019.

\bibitem[Nacson et~al.(2022)Nacson, Ravichandran, Srebro, and
  Soudry]{nacson2022implicit}
Mor~Shpigel Nacson, Kavya Ravichandran, Nathan Srebro, and Daniel Soudry.
\newblock Implicit bias of the step size in linear diagonal neural networks.
\newblock In \emph{International Conference on Machine Learning}, pp.\
  16270--16295. PMLR, 2022.

\bibitem[Paccolat et~al.(2021)Paccolat, Petrini, Geiger, Tyloo, and
  Wyart]{paccolat2021geometric}
Jonas Paccolat, Leonardo Petrini, Mario Geiger, Kevin Tyloo, and Matthieu
  Wyart.
\newblock Geometric compression of invariant manifolds in neural networks.
\newblock \emph{Journal of Statistical Mechanics: Theory and Experiment},
  2021\penalty0 (4):\penalty0 044001, 2021.

\bibitem[Papyan et~al.(2020)Papyan, Han, and Donoho]{papyan2020prevalence}
Vardan Papyan, XY~Han, and David~L Donoho.
\newblock Prevalence of neural collapse during the terminal phase of deep
  learning training.
\newblock \emph{Proceedings of the National Academy of Sciences}, 117\penalty0
  (40):\penalty0 24652--24663, 2020.

\bibitem[Paszke et~al.(2019)Paszke, Gross, Massa, Lerer, Bradbury, Chanan,
  Killeen, Lin, Gimelshein, Antiga, et~al.]{paszke2019pytorch}
Adam Paszke, Sam Gross, Francisco Massa, Adam Lerer, James Bradbury, Gregory
  Chanan, Trevor Killeen, Zeming Lin, Natalia Gimelshein, Luca Antiga, et~al.
\newblock Pytorch: An imperative style, high-performance deep learning library.
\newblock \emph{Advances in neural information processing systems}, 32, 2019.

\bibitem[Payeur et~al.(2021)Payeur, Guerguiev, Zenke, Richards, and
  Naud]{payeur2020burst}
Alexandre Payeur, Jordan Guerguiev, Friedemann Zenke, Blake~A Richards, and
  Richard Naud.
\newblock Burst-dependent synaptic plasticity can coordinate learning in
  hierarchical circuits.
\newblock \emph{Nature neuroscience}, pp.\  1--10, 2021.

\bibitem[Pezeshki et~al.(2021)Pezeshki, Kaba, Bengio, Courville, Precup, and
  Lajoie]{pezeshki2021gradient}
Mohammad Pezeshki, Oumar Kaba, Yoshua Bengio, Aaron~C Courville, Doina Precup,
  and Guillaume Lajoie.
\newblock Gradient starvation: A learning proclivity in neural networks.
\newblock \emph{Advances in Neural Information Processing Systems}, 34, 2021.

\bibitem[Pla{\c{c}}ais \& Preat(2013)Pla{\c{c}}ais and
  Preat]{placcais2013favor}
Pierre-Yves Pla{\c{c}}ais and Thomas Preat.
\newblock To favor survival under food shortage, the brain disables costly
  memory.
\newblock \emph{Science}, 339\penalty0 (6118):\penalty0 440--442, 2013.

\bibitem[Pogodin et~al.(2023)Pogodin, Cornford, Ghosh, Gidel, Lajoie, and
  Richards]{pogodin2023synaptic}
Roman Pogodin, Jonathan Cornford, Arna Ghosh, Gauthier Gidel, Guillaume Lajoie,
  and Blake Richards.
\newblock Synaptic weight distributions depend on the geometry of plasticity.
\newblock \emph{arXiv preprint arXiv:2305.19394}, 2023.

\bibitem[Rajan \& Abbott(2006)Rajan and Abbott]{rajan2006eigenvalue}
Kanaka Rajan and Larry~F Abbott.
\newblock Eigenvalue spectra of random matrices for neural networks.
\newblock \emph{Physical review letters}, 97\penalty0 (18):\penalty0 188104,
  2006.

\bibitem[Raman \& O’Leary(2021)Raman and O’Leary]{raman2021frozen}
Dhruva~V Raman and Timothy O’Leary.
\newblock Frozen algorithms: how the brain's wiring facilitates learning.
\newblock \emph{Current Opinion in Neurobiology}, 67:\penalty0 207--214, 2021.

\bibitem[Richards et~al.(2019)Richards, Lillicrap, Beaudoin, Bengio, Bogacz,
  Christensen, Clopath, Costa, de~Berker, Ganguli, et~al.]{richards2019deep}
Blake~A Richards, Timothy~P Lillicrap, Philippe Beaudoin, Yoshua Bengio, Rafal
  Bogacz, Amelia Christensen, Claudia Clopath, Rui~Ponte Costa, Archy
  de~Berker, Surya Ganguli, et~al.
\newblock A deep learning framework for neuroscience.
\newblock \emph{Nature neuroscience}, 22\penalty0 (11):\penalty0 1761--1770,
  2019.

\bibitem[Roelfsema \& Holtmaat(2018)Roelfsema and
  Holtmaat]{roelfsema2018control}
Pieter~R Roelfsema and Anthony Holtmaat.
\newblock Control of synaptic plasticity in deep cortical networks.
\newblock \emph{Nature Reviews Neuroscience}, 19\penalty0 (3):\penalty0
  166--180, 2018.

\bibitem[Sacramento et~al.(2018)Sacramento, Costa, Bengio, and
  Senn]{sacramento2018dendritic}
Jo{\~a}o Sacramento, Rui~Ponte Costa, Yoshua Bengio, and Walter Senn.
\newblock Dendritic cortical microcircuits approximate the backpropagation
  algorithm.
\newblock \emph{arXiv preprint arXiv:1810.11393}, 2018.

\bibitem[Sadtler et~al.(2014)Sadtler, Quick, Golub, Chase, Ryu, Tyler-Kabara,
  Yu, and Batista]{sadtler2014neural}
Patrick~T Sadtler, Kristin~M Quick, Matthew~D Golub, Steven~M Chase, Stephen~I
  Ryu, Elizabeth~C Tyler-Kabara, Byron~M Yu, and Aaron~P Batista.
\newblock Neural constraints on learning.
\newblock \emph{Nature}, 512\penalty0 (7515):\penalty0 423--426, 2014.

\bibitem[Salaj et~al.(2021)Salaj, Subramoney, Kraisnikovic, Bellec, Legenstein,
  and Maass]{salaj2021spike}
Darjan Salaj, Anand Subramoney, Ceca Kraisnikovic, Guillaume Bellec, Robert
  Legenstein, and Wolfgang Maass.
\newblock Spike frequency adaptation supports network computations on
  temporally dispersed information.
\newblock \emph{Elife}, 10:\penalty0 e65459, 2021.

\bibitem[Saxe et~al.(2013)Saxe, McClelland, and Ganguli]{saxe2013exact}
Andrew~M Saxe, James~L McClelland, and Surya Ganguli.
\newblock Exact solutions to the nonlinear dynamics of learning in deep linear
  neural networks.
\newblock \emph{arXiv preprint arXiv:1312.6120}, 2013.

\bibitem[Saxe et~al.(2019)Saxe, Bansal, Dapello, Advani, Kolchinsky, Tracey,
  and Cox]{saxe2019information}
Andrew~M Saxe, Yamini Bansal, Joel Dapello, Madhu Advani, Artemy Kolchinsky,
  Brendan~D Tracey, and David~D Cox.
\newblock On the information bottleneck theory of deep learning.
\newblock \emph{Journal of Statistical Mechanics: Theory and Experiment},
  2019\penalty0 (12):\penalty0 124020, 2019.

\bibitem[Scellier \& Bengio(2017)Scellier and Bengio]{scellier2017equilibrium}
Benjamin Scellier and Yoshua Bengio.
\newblock Equilibrium propagation: Bridging the gap between energy-based models
  and backpropagation.
\newblock \emph{Frontiers in computational neuroscience}, 11:\penalty0 24,
  2017.

\bibitem[Scheffer et~al.(2020)Scheffer, Xu, Januszewski, Lu, Takemura,
  Hayworth, Huang, Shinomiya, Maitlin-Shepard, Berg,
  et~al.]{scheffer2020connectome}
Louis~K Scheffer, C~Shan Xu, Michal Januszewski, Zhiyuan Lu, Shin-ya Takemura,
  Kenneth~J Hayworth, Gary~B Huang, Kazunori Shinomiya, Jeremy Maitlin-Shepard,
  Stuart Berg, et~al.
\newblock A connectome and analysis of the adult drosophila central brain.
\newblock \emph{Elife}, 9:\penalty0 e57443, 2020.

\bibitem[Schuessler et~al.(2020)Schuessler, Mastrogiuseppe, Dubreuil, Ostojic,
  and Barak]{schuessler2020interplay}
Friedrich Schuessler, Francesca Mastrogiuseppe, Alexis Dubreuil, Srdjan
  Ostojic, and Omri Barak.
\newblock The interplay between randomness and structure during learning in
  rnns.
\newblock \emph{Advances in neural information processing systems},
  33:\penalty0 13352--13362, 2020.

\bibitem[Schuessler et~al.(2023)Schuessler, Mastrogiuseppe, Ostojic, and
  Barak]{schuessler2023aligned}
Friedrich Schuessler, Francesca Mastrogiuseppe, Srdjan Ostojic, and Omri Barak.
\newblock Aligned and oblique dynamics in recurrent neural networks.
\newblock \emph{arXiv preprint arXiv:2307.07654}, 2023.

\bibitem[Seleznova \& Kutyniok(2022)Seleznova and
  Kutyniok]{seleznova2022neural}
Mariia Seleznova and Gitta Kutyniok.
\newblock Neural tangent kernel beyond the infinite-width limit: Effects of
  depth and initialization.
\newblock In \emph{International Conference on Machine Learning}, pp.\
  19522--19560. PMLR, 2022.

\bibitem[Shao \& Ostojic(2023)Shao and Ostojic]{shao2023relating}
Yuxiu Shao and Srdjan Ostojic.
\newblock Relating local connectivity and global dynamics in recurrent
  excitatory-inhibitory networks.
\newblock \emph{PLOS Computational Biology}, 19\penalty0 (1):\penalty0
  e1010855, 2023.

\bibitem[Simard et~al.(2005)Simard, Nadeau, and Kr{\"o}ger]{simard2005fastest}
D~Simard, L~Nadeau, and H~Kr{\"o}ger.
\newblock Fastest learning in small-world neural networks.
\newblock \emph{Physics Letters A}, 336\penalty0 (1):\penalty0 8--15, 2005.

\bibitem[Song et~al.(2016)Song, Yang, and Wang]{song2016training}
H~Francis Song, Guangyu~R Yang, and Xiao-Jing Wang.
\newblock Training excitatory-inhibitory recurrent neural networks for
  cognitive tasks: a simple and flexible framework.
\newblock \emph{PLoS computational biology}, 12\penalty0 (2):\penalty0
  e1004792, 2016.

\bibitem[Song et~al.(2005)Song, Sj{\"o}str{\"o}m, Reigl, Nelson, and
  Chklovskii]{song2005highly}
Sen Song, Per~Jesper Sj{\"o}str{\"o}m, Markus Reigl, Sacha Nelson, and Dmitri~B
  Chklovskii.
\newblock Highly nonrandom features of synaptic connectivity in local cortical
  circuits.
\newblock \emph{PLoS biology}, 3\penalty0 (3):\penalty0 e68, 2005.

\bibitem[Thibeault et~al.(2024)Thibeault, Allard, and
  Desrosiers]{thibeault2024low}
Vincent Thibeault, Antoine Allard, and Patrick Desrosiers.
\newblock The low-rank hypothesis of complex systems.
\newblock \emph{Nature Physics}, pp.\  1--9, 2024.

\bibitem[Tishby \& Zaslavsky(2015)Tishby and Zaslavsky]{tishby2015deep}
Naftali Tishby and Noga Zaslavsky.
\newblock Deep learning and the information bottleneck principle.
\newblock In \emph{2015 ieee information theory workshop (itw)}, pp.\  1--5.
  IEEE, 2015.

\bibitem[Virtanen et~al.(2020)Virtanen, Gommers, Oliphant, Haberland, Reddy,
  Cournapeau, Burovski, Peterson, Weckesser, Bright, et~al.]{virtanen2020scipy}
Pauli Virtanen, Ralf Gommers, Travis~E Oliphant, Matt Haberland, Tyler Reddy,
  David Cournapeau, Evgeni Burovski, Pearu Peterson, Warren Weckesser, Jonathan
  Bright, et~al.
\newblock Scipy 1.0: fundamental algorithms for scientific computing in python.
\newblock \emph{Nature methods}, 17\penalty0 (3):\penalty0 261--272, 2020.

\bibitem[Vodrahalli et~al.(2022)Vodrahalli, Shivanna, Sathiamoorthy, Jain, and
  Chi]{vodrahalli2022nonlinear}
Kiran Vodrahalli, Rakesh Shivanna, Maheswaran Sathiamoorthy, Sagar Jain, and
  Ed~H Chi.
\newblock Nonlinear initialization methods for low-rank neural networks.
\newblock \emph{arXiv preprint arXiv:2202.00834}, 2022.

\bibitem[Winnubst et~al.(2019)Winnubst, Bas, Ferreira, Wu, Economo, Edson,
  Arthur, Bruns, Rokicki, Schauder, et~al.]{winnubst2019reconstruction}
Johan Winnubst, Erhan Bas, Tiago~A Ferreira, Zhuhao Wu, Michael~N Economo,
  Patrick Edson, Ben~J Arthur, Christopher Bruns, Konrad Rokicki, David
  Schauder, et~al.
\newblock Reconstruction of 1,000 projection neurons reveals new cell types and
  organization of long-range connectivity in the mouse brain.
\newblock \emph{Cell}, 179\penalty0 (1):\penalty0 268--281, 2019.

\bibitem[Winston et~al.(2023)Winston, Mastrovito, Shea-Brown, and
  Mihalas]{winston2023heterogeneity}
Chloe~N Winston, Dana Mastrovito, Eric Shea-Brown, and Stefan Mihalas.
\newblock Heterogeneity in neuronal dynamics is learned by gradient descent for
  temporal processing tasks.
\newblock \emph{Neural Computation}, 35\penalty0 (4):\penalty0 555--592, 2023.

\bibitem[Woodworth et~al.(2020)Woodworth, Gunasekar, Lee, Moroshko, Savarese,
  Golan, Soudry, and Srebro]{woodworth2020kernel}
Blake Woodworth, Suriya Gunasekar, Jason~D Lee, Edward Moroshko, Pedro
  Savarese, Itay Golan, Daniel Soudry, and Nathan Srebro.
\newblock Kernel and rich regimes in overparametrized models.
\newblock In \emph{Conference on Learning Theory}, pp.\  3635--3673. PMLR,
  2020.

\bibitem[Xiao et~al.(2020)Xiao, Pennington, and
  Schoenholz]{xiao2020disentangling}
Lechao Xiao, Jeffrey Pennington, and Samuel Schoenholz.
\newblock Disentangling trainability and generalization in deep neural
  networks.
\newblock In \emph{International Conference on Machine Learning}, pp.\
  10462--10472. PMLR, 2020.

\bibitem[Xie et~al.(2022)Xie, Muscinelli, Harris, and
  Litwin-Kumar]{xie2022task}
Marjorie Xie, Samuel Muscinelli, Kameron~Decker Harris, and Ashok Litwin-Kumar.
\newblock Task-dependent optimal representations for cerebellar learning.
\newblock \emph{bioRxiv}, pp.\  2022--08, 2022.

\bibitem[Yang(2020)]{yang2020tensor}
Greg Yang.
\newblock Tensor programs ii: Neural tangent kernel for any architecture.
\newblock \emph{arXiv preprint arXiv:2006.14548}, 2020.

\bibitem[Yang \& Molano-Maz{\'o}n(2021)Yang and
  Molano-Maz{\'o}n]{yang2021towards}
Guangyu~Robert Yang and Manuel Molano-Maz{\'o}n.
\newblock Towards the next generation of recurrent network models for cognitive
  neuroscience.
\newblock \emph{Current opinion in neurobiology}, 70:\penalty0 182--192, 2021.

\bibitem[Yang \& Wang(2020)Yang and Wang]{yang2020artificial}
Guangyu~Robert Yang and Xiao-Jing Wang.
\newblock Artificial neural networks for neuroscientists: a primer.
\newblock \emph{Neuron}, 107\penalty0 (6):\penalty0 1048--1070, 2020.

\bibitem[Zador(2019)]{zador2019critique}
Anthony~M Zador.
\newblock A critique of pure learning and what artificial neural networks can
  learn from animal brains.
\newblock \emph{Nature communications}, 10\penalty0 (1):\penalty0 1--7, 2019.

\bibitem[Zhao et~al.(2011)Zhao, Beverlin, Netoff, and
  Nykamp]{zhao2011synchronization}
Liqiong Zhao, Bryce Beverlin, Theoden Netoff, and Duane~Q Nykamp.
\newblock Synchronization from second order network connectivity statistics.
\newblock \emph{Frontiers in computational neuroscience}, 5:\penalty0 28, 2011.

\end{thebibliography}
\bibliographystyle{iclr2024_conference}

\newpage

\appendix

\section{Extended discussions on related works} \label{scn:related_works}

\textbf{Theoretical Foundations of Neural Network Regimes and Implications for Neural Circuits:} The journey of understanding deep learning systems has borne witness to unprecedented progress in the mathematical dissection of neural network functionalities \citep{advani2020high,jacot2018neural,pezeshki2021gradient,baratin2021implicit,alemohammad2020recurrent,yang2020tensor,agarwala2022second,atanasov2021neural,azulay2021implicit,emami2021implicit}. These theoretical findings, until recently confined predominantly to artificial domains, have embarked upon explorations into biological neural networks, elucidating the intricate dynamics of learning and computational properties \citep{bordelon2022influence,liu2022beyond,braun2022exact,ghosh2023gradient}. Among the vanguard of these theoretical endeavors stands the dichotomy of 'rich' and 'lazy' learning regimes. Both lead to task learning, yet they carry distinct ramifications for representation and generalization \citep{chizat2019lazy,flesch2021rich,geiger2020disentangling,george2022lazy,ghorbani2020neural,woodworth2020kernel,paccolat2021geometric,nacson2022implicit,haochen2021shape,flesch2023continual}. In the 'lazy' regime, which is typically associated with large initial weights, learning predominantly centers on adjusting the readout weights. This leads to minimal alterations in the network weights and representation, while capitalizing on the expansive dimensionality provided by the hidden layer's random projections~\citep{flesch2021rich}. In contrast, the 'rich' regime, defined by smaller initial weights, fosters the development of highly tailored hidden unit representations specifically aligned with task demands, resulting in considerable adaptations in weights and representation. It's essential to highlight that the transition and dominance between these regimes are influenced by more than just the initial weight scale. Other factors, ranging from network width to the output gain (often referred to as the $\alpha$ parameter), play a pivotal role \citep{chizat2019lazy,geiger2020disentangling}. 

A nexus between deep learning theoretical frameworks and neuroscience has unveiled applications of the rich/lazy regimes. Previous investigations characterized neural network behaviors under distinct regimes \citep{bordelon2022influence,schuessler2023aligned} and discerning which mode yields solutions mimicking empirical data \citep{flesch2021rich}. It is compelling to observe that the existence of multiple learning regimes isn't an isolated phenomenon in artificial systems; analogous learning patterns echo in neural circuits as well. For instance, while plasticity-driven transformations might be resource-intensive, they manifest robustly during such developmental phases, followed by minor changes afterwards \citep{lohmann2014developmental}. Building upon these findings, our research delves deeper into the precursors of these regimes. We examine how inherent factors in the brain, especially initial weight configurations, influence the inclination towards either rich or lazy learning. This understanding is crucial for assessing the applicability of regime-specific tools in neural contexts and for shedding light on the potential benefits of having both learning regimes coexist in the brain.

\textbf{Interplay of Neural Learning and structure:} Understanding how the brain learns using its myriad elements is a perennial quest in neuroscience. Addressing this, certain studies have unveiled biologically plausible learning rules \citep{lillicrap2020backpropagation,scellier2017equilibrium,diederich1987learning,hinton2022forward,laborieux2022holomorphic,greedy2022single,sacramento2018dendritic,payeur2020burst,roelfsema2018control,meulemans2022least,murray2019local,bellec2020solution,liu2021cell,liu2022biologically,marschall2020unified}, suggesting potential neural algorithms involving known neural ingredients. Concurrently, given the three primary components of a neural network's design --- task, learning rule, and architecture --- another avenue of research delves deep into the architectural facet, specifically focusing on how it interacts with the learning rule to enhance learning \citep{richards2019deep,zador2019critique,yang2021towards}. Under the structural umbrella, the neural unit's complexity and initial connectivity patterns are two crucial aspects. Complex neuron models, for instance, have shown the potential in boosting learning performance by allowing implicit forms of memory and computations at the single neuron level~\citep{salaj2021spike,winston2023heterogeneity}. Moreover, A large body of work has investigated the effect of different random initializations on learning in deep networks \citep{saxe2013exact,bahri2020statistical,glorot2010understanding,he2015delving,arora2019theory}. For instance, the variance in random initial weights can induce pronounced shifts in network behavior, ranging from the "lazy" to the "rich" regimes \citep{chizat2019lazy,flesch2021rich}. This introduces unique inductive biases during the learning process, with distinct preferences for learning certain features \citep{george2022lazy}. Our discourse primarily orbits around connectivity and its implications on learning dynamics in networks with simple rectified units. Our results sit within the purview of these regimes, with a widely adopted assumption of gradient descent via backpropagation as the learning rule, while remaining open to encompassing a wider spectrum of rules in future explorations.

\textbf{Neural circuit connectivity pattern and eigenspectrum:} While the importance of initial weights on function and learning is clear, the impact of specific weight shapes, apart from weight scale, on rich or lazy dynamics remains less explored. The predominant focus in the literature has been on random initialization. Yet, neural circuit structures significantly diverge from this paradigm. Illustratively, one finds connectivity principles or patterns markedly different from what one observes with a mere random initialization \citep{pogodin2023synaptic}, resulting in distinct neural dynamics; these connectivity principles or patterns include Dale's law \citep{rajan2006eigenvalue,ipsen2020consequences,harris2022eigenvalue}, an over-representation of higher-order motifs \citep{dahmen2020strong} and cell-type-specific connectivity statistics \citep{aljadeff2015transition}, to name a few. Given the prominence of low-rankedness observed in neural circuits \citep{song2005highly}, our study centers on the influence of effective rank on the effective learning regime. As the next generation of connectivity data becomes available \citep{campagnola2022local,microns2021functional,dorkenwald2022flywire,winnubst2019reconstruction,scheffer2020connectome}, future explorations will broaden the scope to other weight characteristics.

\newpage 

\section{Proofs} \label{scn:proofs}

\subsection{Proofs for main text Theorem and Proposition}

\paragraph{Notation}  Let $f(x) = W_2 W_1 x$ denote a two-layer linear network with $N$ hidden units on $d$-dimensional inputs $x\in \R^d$, with weight matrices $ W_1 \in \R^{N \times d}$ and $W_2 \in \R^{1 \times N}$. 
We consider $m$ training inputs $x_1, \cdots x_m$  and the corresponding data matrix $X = [x^T_1 \cdots x^T_m] \in \R^{d \times m}$ ;  the output target is generated from a linear teacher network as $Y = \beta^T X$, where $\beta_i \sim  \mathcal{N}(0, 1/d)$. 

Since our goal is to investigate how the {\it shape} of the initial weights impacts network change, we will consider a fixed small (Froebenius) norm for these; i.e.,  
\begin{equation*}
\| W_1^{(0)} \|_F = \|W_2^{(0)} \|_F := \sigma \ll 1
\end{equation*} 
We denote by $s_1, \cdots,  s_d$ denote the singular values of $W_1^{(0)}$; they satisfy  $\sum_{j=1}^d s^2_j  = \sigma^2$. 

In what follows we focus on the whitened setting, where $X$ has all its non zero singular values equal to 1.   We also  assume $m \geq d$ for simplicity (this assumption can easily be relaxed in our analysis), so that the whitened data assumption translates as  $XX^T = I_d$.

\paragraph{Prior results} Our analysis builds on prior results  \citet{atanasov2021neural}  on the evolution of the NTK for two-layer linear networks trained by gradient flow of the mean square error.  In the above setting, 
\citet{atanasov2021neural} show that the final NTK $K^{(f)}$ (i.e. the asymptotic NTK as the number of iterations goes to infinity) is given by
\begin{align} 
   K^{(f)} = ||\beta||  X^T (\hat{\beta} \hat{\beta}^T + I_d) X + O(\sigma^2). \label{eqn:NTKf}
\end{align}
where $\hat{\beta}:=\beta / \|\beta\|$. 
We are interested in the expected kernel alignment over the tasks, in the small initialization regime: 
\begin{align} \label{eqn:NTK_tilde_align}
\mathbb{E}_\beta [{KA} (K^{(f)}, K^{(0)})] &:= \mathbb{E}_\beta  \left[\frac{\Tr(K^{(f)} K^{(0)})}{\|K^{(f)} \|_F \| K^{(0)}\|_F}\right].
\end{align}

\addtocounter{thrm}{-1} 
\begin{thrm}  In the above setting, 
 when considering all possible initializations $W_1^{(0)}$ with small fixed norm $\sigma$, the expected kernel alignment  $\mathbb{E}_\beta[KA]$ (defined in Eq.~\ref{eqn:NTK_tilde_align}) is maximized with high-rank isotropic initialization, i.e with $W_1^{(0)}$ that has all its non-zero singular values equal in absolute value.   
\end{thrm}
\begin{proof} Let us write $K^{(0)}=X^T M_0 X$ with $M_0:=W^{(0) T}_1 W^{(0)}_1 + \sigma^2 I_d$. Up to $O(\sigma^4)$ terms,  the numerator  in Eq.~\ref{eqn:NTK_tilde_align} takes the form 
\begin{align} \label{eq:num} 
\Tr(K^{(f)} K^{(0)}) &=  \| \beta \| \Tr (X^T (\hat{\beta} \hat{\beta}^T + I_d) X X^T M_0 X) \cr 
&\overset{(a)}{=}  \| \beta \|  \Tr(X^T (\hat{\beta} \hat{\beta}^T + I_d) M_0 X) \cr 
&\overset{(b)}{=}  \| \beta \|  \Tr((\hat{\beta} \hat{\beta}^T + I_d) M_0 XX^T) \cr 
&\overset{(a)}{=} \| \beta \|  \Tr ((\hat{\beta} \hat{\beta}^T + I_d) M_0) \cr 
&\overset{(c)}{=} \| \beta \|  (\hat\beta^T M_0 \hat\beta  + \Tr  M_0)  
\end{align}
where $(a)$ uses $XX^T = I_d$, $(b)$ the cyclicity of the trace, and $(c)$ the fact that $\hat\beta^T M_0 \hat\beta$ is a scalar. 

As for the denominator in Eq.~\ref{eqn:NTK_tilde_align}), we have,  
\begin{align} \label{eq:den1} 
    \|K^{(0)} \|^2_F &= \Tr(K^{{(0)} } K^{(0)}) \cr 
     &= \Tr(X^T M_0 X X^T M_0 X) \cr 
     &\overset{(a)}{=} \Tr(M_0^2) \cr 
    \end{align}
 and, up to $O(\sigma^4)$ terms, 
    \begin{align} \label{eq:den2}  
    \| K^{(f)} \|_F^2 &= \Tr(K^{{(f)} } K^{(f)}) \cr 
    &= \| \beta\| ^2 \Tr (X^T (\hat\beta \hat\beta^T + I_d) XX^T (\hat\beta \hat\beta^T + I_d) ) \cr 
    &= \| \beta\| ^2\Tr (X^T (\hat\beta \hat\beta^T + I_d) X ) \cr 
    &\overset{(a)}{=}  \| \beta\| ^2 \Tr (\hat\beta \hat\beta^T + I_d)^2 \cr 
    &\overset{(b)}{=} \| \beta\| ^2 (d + 3)
    \end{align} 
where $(a)$ in these two calculations uses $XX^T  =I_d$ and the cyclicity of the trace;  and $(b)$ notes that the $d \times d$   matrix $\hat\beta \hat\beta^T + I_d$ has $d-1$ eigenvalues equal to 1 and one equal to $2$.  
Eq. \ref{eq:den1} and \ref{eq:den2} yield
\begin{equation} \label{eq:denom}
\| K^{(f)} \|_F   \|K^{(0)} \|_F = \|\beta\| \sqrt{(d+3) \Tr M_0^2}
\end{equation} 
Putting together Eq.~\ref{eq:num}, \ref{eq:denom} , we obtain, up to additive $O(\sigma^2)$ terms, 
\begin{equation}
\mathrm{KA}(K^{(f)}, K^{(0)}) = \frac{\hat\beta^T M_0 \hat\beta  + \Tr M_0}{\sqrt{(d+3) \Tr M_0^2}} 
\end{equation}
Next, averaging over the tasks requires computing the Gaussian average 
\begin{equation*} 
A[M_0] := \E_\beta \left[ \hat\beta^T  M_0 \hat\beta \right] = \E_\beta \left[ \frac{\beta^T M_0 \beta}{\|\beta\|^2}\right] . 
  \end{equation*}
\begin{lemma} \label{appendlem_inv}
The map $A$ 
 is invariant under the action of the orthogonal group, i.e  $A[U\! M \! U^T] = A[M]$ for all $M \in \R^{d\times d}$ and all orthogonal matrices $U \in \R^{d\times d}$.  
\end{lemma}
\begin{proof} 
This is a consequence of the invariance of the Gaussian measure under the action of the orthogonal group. Explicitly, given an orthogonal matrix $U$, 
\begin{align} 
A[U\! M \! U^T] &= \frac{1}{(2\pi d)^{d/2}} \int \mathrm{d}^d\beta  e^{-\|\beta\|^2 /d} \left[ \frac{\beta^T U M U^\top \beta}{\|\beta\|^2}\right] \cr
&\overset{\beta' := U^T \beta}{=} \frac{1}{(2\pi d)^{d/2}}  \int \mathrm{d}^d\beta'  |\mathrm{det} U| e^{-\|U\beta'\|^2 /d} \left[ \frac{\beta'^T M \beta'}{\|U\beta'\|^2}\right] \cr
&= \frac{1}{(2\pi d)^{d/2}}  \int \mathrm{d}^d\beta'  e^{-\|\beta'\|^2 /d} \left[ \frac{\beta'^T M \beta'}{\|\beta'\|^2}\right] \cr
&= A[M]
\end{align} 
where the third equality follows from $|\det U | = 1$ and $\|U\beta \| = \|\beta \|$. 
\end{proof} 
\begin{lemma}
There is a constant $c$ such that $A[M] = $c$ \Tr(M)$  for any symmetric matrix $M$. 
\end{lemma}
\begin{proof} 
Given a symmetric matrix $M$, it can be diagonalized as  $M = U D U^T$ where $D = \mathrm{Diag}(\mu_1, \cdots \mu_d)$ is diagonal and  $U$ is orthogonal. By rotation invariance from Lemma \ref{appendlem_inv}, we have $A[M] = A[D]$, and 
\begin{equation} 
A[D] = \E_\beta \left[ \hat\beta^T D \hat\beta \right]  
=  \E_{\beta} \left[ \sum_{j=1}^d \hat \beta_j^2 \mu_j \right] 
= \sum_{j=1}^d \E_{\beta} \left[\frac{\beta_j^2}{\| \beta \|^2} \right]  \mu_j := \sum_{j=1}^d c_j  \mu_j
\end{equation} 
We conclude by noting that, by invariance of the (isotropic) Gaussian measure under permutation of the vector components,   the coefficients $c_j $ are independent of $j$,  i.e $c_j \equiv c$ for all $j$. In sum,  
 \begin{equation} 
 A[M] = A[D] = c \Tr D = c \Tr M.
 \end{equation}
\end{proof}
The expected kernel alignment thus takes the form, 
\begin{equation} \label{eq:proofEKA}
\E_{\beta} [\mathrm{KA}(K^{(f)}, K^{(0)})] = \frac{(1+c) \Tr M_0}{\sqrt{(d+3) \Tr M_0^2}} 
\end{equation} 
up to additive $O(\sigma^2)$ terms. 
Finally, we note that  
\begin{align} 
\Tr M_0 &= \Tr(W^{(0) T}_1 W^{(0)}_1 + \sigma^2 I_d) \cr
&=  \| W^{(0)}_1\|^2_F + d\sigma^2  \cr
 &= (d+1)\sigma^2
\end{align} 
and 
\begin{align} 
\Tr M^2_0  &= \Tr (W^{(0) T}_1 W^{(0)}_1 + \sigma I_d)^2 \cr
&=  \sum_{j=1}^d (s_j^2 + \sigma^2)^2 \cr
&= \sum_{j=1}^d s_j^4 + 2 \sigma^2 \sum_{j=1}^d s_j^2 + d\sigma^4 \cr
&= \sum_{j=1}^d s^4_j + (d+2) \sigma^4 
\end{align} 
Substituting into Eq.~\ref{eq:proofEKA}, we have, up to additive $O(\sigma^2)$ terms, 
\begin{equation} \label{eq:finalEKA}
    \E_{\beta} [\mathrm{KA}(K^{(f)}, K^{(0)})]  = \frac{(1+c) (d+1)}{\sqrt{(d+3)(d+2 + \sum_{j=1}^d (s_j/\sigma)^4)}} 
\end{equation}

Finally, we see in Eq \ref{eq:finalEKA}  that the maximization of $\mathbb{E}_\beta[KA]$ reduces to the following convex constrained optimization problem:
\begin{equation}
    \min_{s} \sum_j s_j^4, \quad \text{subject to } \sum_j s_j^2 = \sigma^2.
\end{equation}
The KKT solutions satisfy  $s^2_i  = \sigma^2 /d$ for all $j = 1\cdots d$.   
This implies that the expected tangent kernel alignment is maximized when the initial weight singular values $|s_i|$ are distributed evenly across dimensions, which corresponds to a high-rank initialization.
\end{proof}

\addtocounter{prop}{-1}
\begin{prop} 
 Following the setup and assumptions in Theorem~\ref{thm:NTKalign}, rank-1 initialization with $W^{(0)}_1 = \sigma [\hat\beta^T \quad \vec{0} \quad ... \quad \vec{0}]$ leads to maximal aligment, i.e, 
 $\mathrm{KA}(K^{(f)}, K^{(0)}) = 1$ up to additive $O(\sigma^2)$ terms. 
\end{prop}

\begin{proof} 
We indeed have, 
\begin{align}
    K^{(0)} &= X^T (W_1^{{(0)}^T} W^{(0)}_1 + \| W^{(0)}_2 \|^2 I ) X \cr
    &= \sigma^2 X^T (\hat\beta \hat\beta^T + I ) X 
\end{align}
Thus, writing $K:=X^T (\hat\beta \hat\beta^T + I ) X$ and using Eq. \ref{eqn:NTKf}, the alignment takes the form
\begin{align}
\mathrm{KA}(K^{(f)}, K^{(0)}) &:= \frac{\Tr(K^{(f)} K^{(0)})}{\|K^{(f)} \|_F \| K^{(0)}\|_F} \cr
&= \frac{\Tr(K (K+O(\sigma^2))}{|K\|_F \|K +O(\sigma^2)\|_F} \cr
&= \frac{\Tr(K^2)}{\|K\|^2_F} +O(\sigma^2)\cr
&= 1 + O(\sigma^2)
\end{align} 

\end{proof}

\subsection{Learning requirement based on $W_h^{(0)}$ rank}

The focus of this idea is to show that no changes to hidden weights $W_h$ is not possible (e.g. reservoir settings) for zero-error when the initial weight rank falls below a certain threshold. Freezing the hidden weights $W_h$ would be a special case of lazy learning. 

\begin{prop} \label{prop:change_requirement}
    Consider a linear RNN with input at time $t$ as $X_t \in \mathbb{R}^{N \times d}$ (for $t = 1, ... , T-1$), target output $Y \in \mathbb{R}^{N_{out} \times d}$ only at the last step, recurrent weight matrix $W_h \in \mathbb{R}^{N \times N}$ and readout weight matrix $w \in \mathbb{R}^{N_{out} \times N}$. Here, $N, N_{out}, d$ and $T$ are the number of hidden units, number of classes, number of data points and number of time steps, respectively, and we assume $N, d > N_{out}$. Define initial recurrent weight $W_h^{(0)}$ and final recurrent weight $W_h^{(f)}$ that achieves zero error. Then, for arbitrary input $X$ and target output $Y$, $W_h^{(f)} = W_h^{(0)}$ is not possible when $rank(W_h^{(0)}) < N_{out}$. 
\end{prop}

\begin{proof}

    We have the following based on the assumption of the RNN structure, if zero-error learning is achieved: 

    \begin{equation}
        Y = w^{(f)} W_h^{(f)} \left(\sum_{t=1}^{T-1} W_h^{{(f)}^{T-t-1}} X_t \right).
    \end{equation}
    
    We can prove by contradiction. Suppose $W_h^{(f)} = W_h^{(0)}$, then

    \begin{equation}
        Y = w^{(f)} W_h^{(0)} \left(\sum_{t=1}^{T-1} W_h^{(0)}{^{T-t-1}} X_t \right).
    \end{equation}
    
    Since $Y$ is arbitrary, we can have $rank(Y) = N_{out}$ (by the assumption of $N, d > N_{out}$). Applying $rank(W_h^{(0)}) < N_{out}$ we have

    \begin{align}
        rank(Y) &= rank(w^{(f)} W_h^{(0)} \left(\sum_{t=1}^{T-1} W_h^{{(0)}^{T-t-1}} X_t \right)) \cr 
        & \overset{(a)}{\leq} min(rank(w^{(f)}), rank(W_h^{(0)}), \left(\sum_{t=1}^{T-1} W_h^{{(0)}^{T-t-1}} X_t \right)) \cr
        & < N_{out},
    \end{align}
    where $(a)$ is because $rank(W_h^{(0)})<N_{out}$ so the minimum has to be less than $N_{out}$. This would contradict an arbitrary $Y$ with $rank(Y) = N_{out}$. Thus, $W_h^{(f)} = W_h^{(0)}$ cannot happen and recurrent weights have to be adjusted. 
\end{proof}

\newpage

\section{Setup and simulation details} \label{scn:sim_details}

\subsection{Initial low-rank weights creation}

For the null case, we initialized with random Gaussian distributions where each weight element \(W_{ij} \sim \mathcal{N}(0, g^2/N)\), with an initial weight variance of \(g\). Unless otherwise mentioned, we set \(g=1.5\) and network size \(N=300\), though we also validated across other parameter choices (see Appendix~\ref{scn:more_sims}). Input and readout weights were initialized similarly as in \citet{yang2020artificial} (see their \(EIRNN.ipynb\) notebook). 

To create low-rank weights using SVD, we generated temporary weights \(\hat{W}_{ij} \sim \mathcal{N}(0, g^2/N)\). Subsequently, we applied SVD to \(\hat{W}\) and retained the top components based on the desired rank. To ensure comparisons are made across constant initial weight magnitudes, the resultant weight matrix was rescaled to match the Frobenius norm of \(\hat{W}\).

Furthermore, we present details for experimentally-driven low-rank weights. For block-specific statistics, we followed the setup in Figure S3 of \citet{aljadeff2015transition}, setting parameters as \(\alpha=0.02\), \(\gamma=10\), and \(1-\epsilon=0.8\). These parameters substantially influence the weight eigenspectrum, as depicted in Figure S3 of \citet{aljadeff2015transition}; we selected these values specifically to emphasize the outliers and achieve a lower effective rank. These parameters represent the fraction of hyperexcitable neurons (population 1), gain of hyperexcitable connections, and the gain of remaining connections, respectively. For the creation of a chain motif, we employed the procedure described in Section S3.10 of \citet{dahmen2020strong}, setting \(\tau_{chn}=0.03\) (and \(\tau_{chn}=-0.1\) for over-representation or under-representation of the chain motif, respectively). Here, we set \(N=100\). These parameters were chosen to provide enough distinctions from the null case, while still ensuring stability and effective task learning. The electron microscopy (EM) connectivity (of the V1 cortical column model) is obtained from \citet{alleninstitute2023brain}, which includes dendritic tree reconstructions and local axonal projections for hundreds of thousands of neurons, detailing their 0.5 billion synaptic connections. From this, we selected 198 cells, focusing on fully proofread neurons closest to the midpoint between layers 2/3 and 4. Connectivity strength for each neuron is determined by summing the volume of each post-synaptic density to target cells, distinguishing between excitatory and inhibitory cell types. For instance, if cell 'a' forms 10 synapses with cell 'b', the connection strength of connection[a,b] represents the combined volume of synaptic densities at cell 'b'. Inhibitory connections are assigned a sign of -1, while excitatory ones receive +1. For the Dale's law obeying initial connectivity, balanced initialization was done following the process in \citet{yang2020artificial} with 80\% excitatory and 20\% inhibitory neurons (see the notebook \(EIRNN.ipynb\)).

It is  crucial to highlight that, in testing our Theorem, which examines the effect of the \textbf{initial} weight rank, all low-rank modifications are not enforced during training (although the impact of enforcing these structures could be an interesting avenue for future exploration). Weights are adjusted freely based on gradient descent learning. 

\subsection{Task and training details}

Our code is accessible at \url{https://github.com/Helena-Yuhan-Liu/BioRNN_RichLazy}. We used PyTorch Version 1.10.2 \citep{paszke2019pytorch}. Simulations were executed on a computer server with x2 20-core Intel(R) Xeon(R) CPU E5-2698 v4 at 2.20GHz, with the average task training duration being around 10 minutes. Following the procedure in \citet{george2022lazy}, which delved deeply into effective laziness metrics, we employed gradient-descent learning with the SGD optimizer. Unless mentioned otherwise, the learning rate was $3e-3$, but we validated that our findings remain consistent across various learning rates (see Appendix~\ref{scn:more_sims}). For stopping, we trained the neurogym tasks for 10000 SGD iterations, which led to comparable terminal losses and accuracies across initializations. For the sMNIST task, we concluded our training upon reaching 97\% accuracy, a criterion informed by both published results and our computational resources. We also experimented with halving and doubling the training iterations and observed similar trends. All weights --- input, recurrent and readout --- were trained.  For statistical analysis and significance tests, we used methods in the SciPy Package \citep{virtanen2020scipy}.  

For the neuroscience tasks, we adopted the Neurogym framework \citep{molano2022neurogym}. Within this paper, these tasks are denoted as "2AF", "DMS", and "CXT", mirroring Neurogym settings: $task='PerceptualDecisionMaking-v0'$, $task='DelayMatchSample-v0'$, and $task='ContextDecisionMaking-v0'$, respectively. To expedite simulations and facilitate numerous runs, we operated with $dt=\tau_m=100ms$ and abbreviated task durations: for 2AF, settings were $stimulus=700ms$ and $decision=100ms$; for DMS, they were $sample=100ms$, $delay=500ms$, $test=100ms$, and $decision=100ms$; for CXT, they comprised $stimulus=200ms$, $delay=500ms$, and $decision=100ms$. For these three tasks, we used a batch size of 32 and trained for 10000 iterations.  

Regarding the sequential MNIST task~\cite{lecun1998mnist}, we employed a row-by-row format to hasten simulations. Inputs were delivered via $N_{in}=28$ units, each presenting a row's grey-scaled value, culminating in 28 steps with network predictions rendered at the final step. Training hinged on the cross-entropy loss function; targets were provided throughout training for the neuroscience tasks, as per Neurogym implementation, and targets were provided at only a trial's conclusion for the sequential MNIST task. For this task, we used a batch size of 200 and trained for 10000 iterations. 

For the student-teacher two-layer linear network simulations in Figure~\ref{fig:linear2layer}A and Figure~\ref{fig:aligned}, we set $N=1000$, $d = 2$ (also found similar trends for $d=20$ and $d=100$), $z=Fx$, all entries of $w$ (or $\beta$ when $F=I$) to $1$ and entries of \( X \) are sampled from a uniform distribution over the interval \([-2, 2]\). We used standard Normal initialization for both $W_1$ and $W_2$ with $\sigma=0.001$. For Figure~\ref{fig:aligned}, $F$ is constructed from SVD, i.e. $F=USV^T$, with $U$ and $V$ generated from arbitrary orthogonal matrices, and $S$ is a diagonal matrix consisting of the singular values with the top half of the singular values set to $\kappa$ and bottom half set to $1$, where $\kappa$ is the condition number of $F$. For the aligned initialization, $W_1$ is initialized as given in Proposition~\ref{prop:aligned_ini} with $\beta = w^T F$ ($w$ here is illustrated in Figure~\ref{fig:aligned}), and the $F$ is replaced by its rank-$(d/2)$ truncation for the partially aligned initialization case. For the MNIST task shown in Figure~\ref{fig:linear2layer}B, we used a two-layer feedforward network with a ReLU activation function. The architecture consists of an input layer with 784 units corresponding to the image pixels, a hidden layer with 300 units, and a linear readout layer with 10 output units. The weights of the hidden layer and the readout layer were initialized similarly to the input and readout weights, respectively, used in the RNN settings.

\newpage  

\section{Additional simulations} \label{scn:more_sims}

\begin{figure}[h!]
    \centering
    \includegraphics[width=0.99\textwidth]{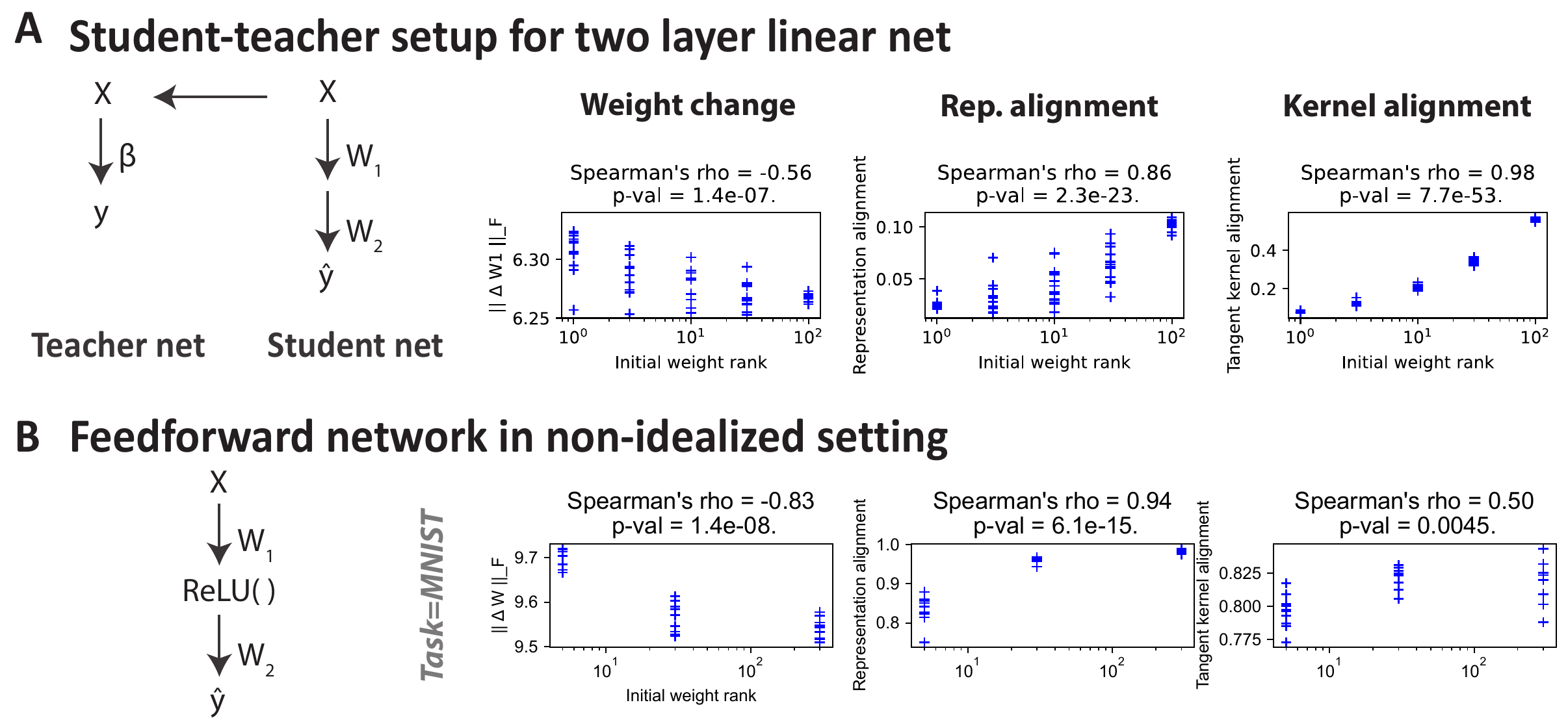}
    \caption{\textbf{As predicted by the theoretical results, higher rank random initialization leads to effectively lazier learning in two-layer linear network. } A) We use the student-teacher two-layer linear network setup described in Section~\ref{scn:theory}. B) a non-idealized setting: two-layer feedforward network with ReLU activation and 300 hidden units trained on the MNIST dataset. Plotting convention follows that of Figure~\ref{fig:svd}. 
    } 
    \label{fig:linear2layer}
\end{figure}

\begin{figure}[h!]
    \centering
    \includegraphics[width=0.99\textwidth]{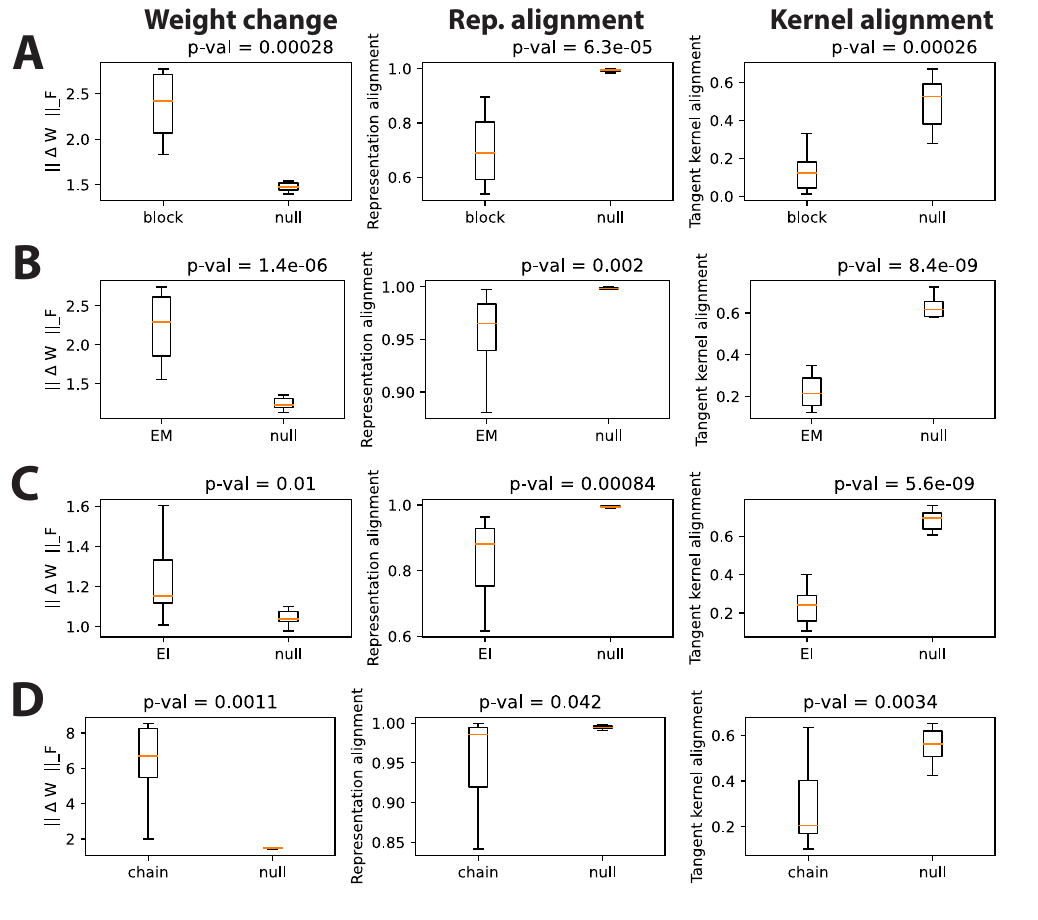}
    \caption{ We repeated Figure~\ref{fig:bio_spectrum_laziness} for the DMS task and observed similar trends: low-rank initialization, achieved by experimentally-driven initial connectivity in Figure~\ref{fig:bio_spectrum_laziness}, leads to effectively richer learning. The plotting conventions used here follow those in Figure~\ref{fig:bio_spectrum_laziness}, with panels A-D corresponding to the ones in that figure.
    } 
    \label{fig:bio_laziness_DMS}
\end{figure}

\begin{figure}[h!]
    \centering
    \includegraphics[width=0.99\textwidth]{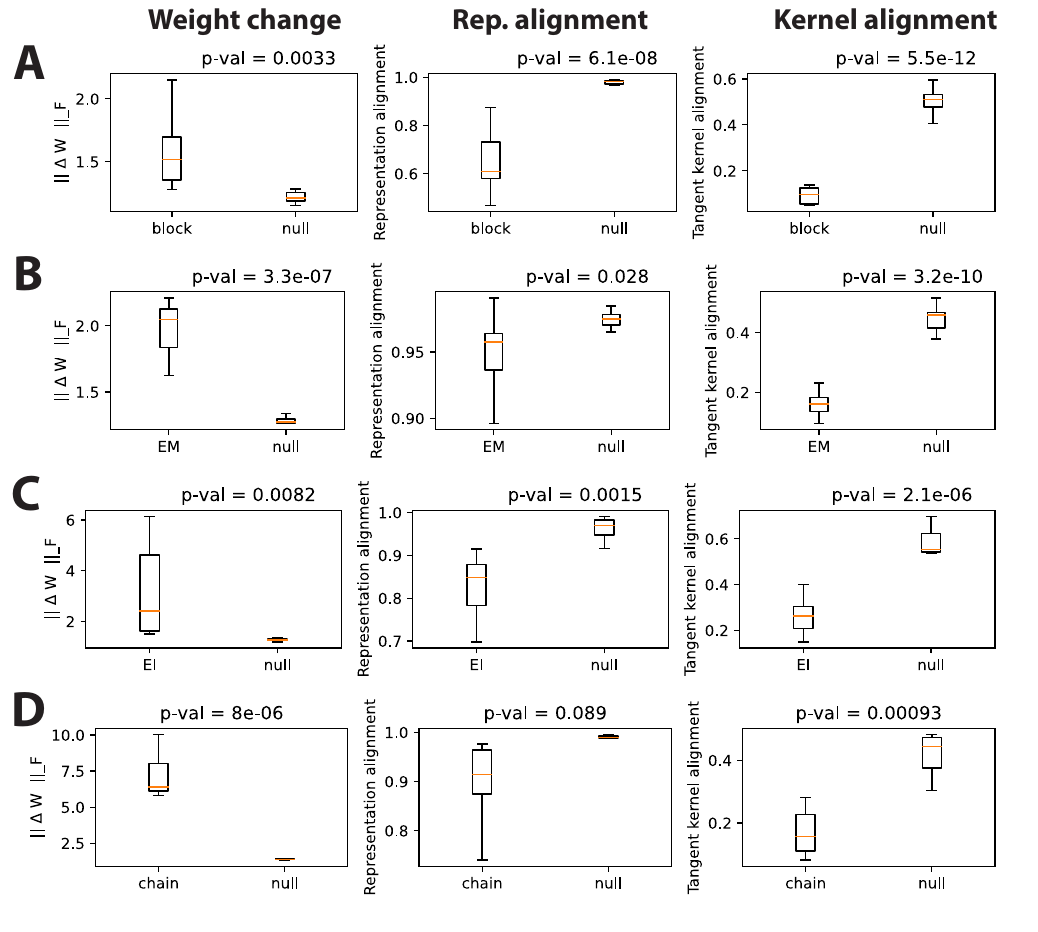}
    \caption{ We repeated Figure~\ref{fig:bio_spectrum_laziness} for the CXT task and observed similar trends: low-rank initialization, achieved by experimentally-driven initial connectivity in Figure~\ref{fig:bio_spectrum_laziness}, leads to effectively richer learning. The plotting conventions used here follow those in Figure~\ref{fig:bio_spectrum_laziness}, with panels A-D corresponding to the ones in that figure. 
    } 
    \label{fig:bio_laziness_CXT}
\end{figure}

\begin{figure}[h!]
    \centering
    \includegraphics[width=0.99\textwidth]{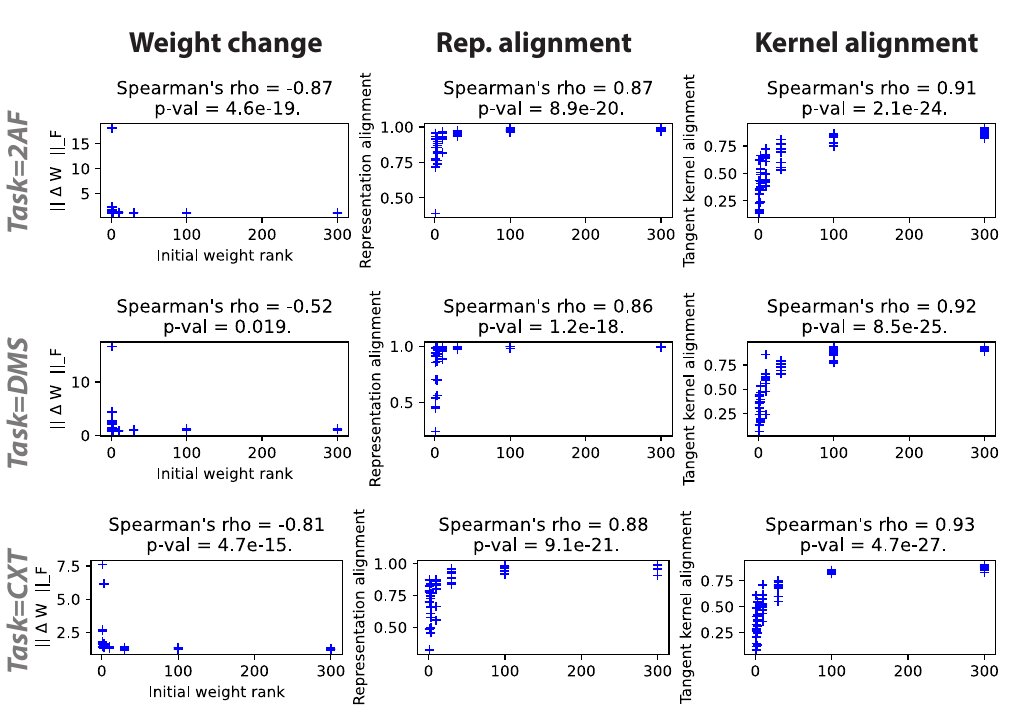}
    \caption{ \textbf{Consistent trends observed in Figure~\ref{fig:svd} also for Uniform initialization}. We replicated the results of Figure~\ref{fig:svd} --- where the initial weights follow a zero-mean Gaussian distribution \(W_{ij} \sim \mathcal{N}(0, g^2/N)\) --- but now for Uniform initialization $W_{ij} \sim \mathcal{U}\left(-\frac{g}{\sqrt{N}}, \frac{g}{\sqrt{N}}\right)$. Plotting conventions follow that of Figure~\ref{fig:svd}.
    } 
    \label{fig:svd_uniform}
\end{figure}

\begin{figure}[h!]
    \centering
    \includegraphics[width=0.99\textwidth]{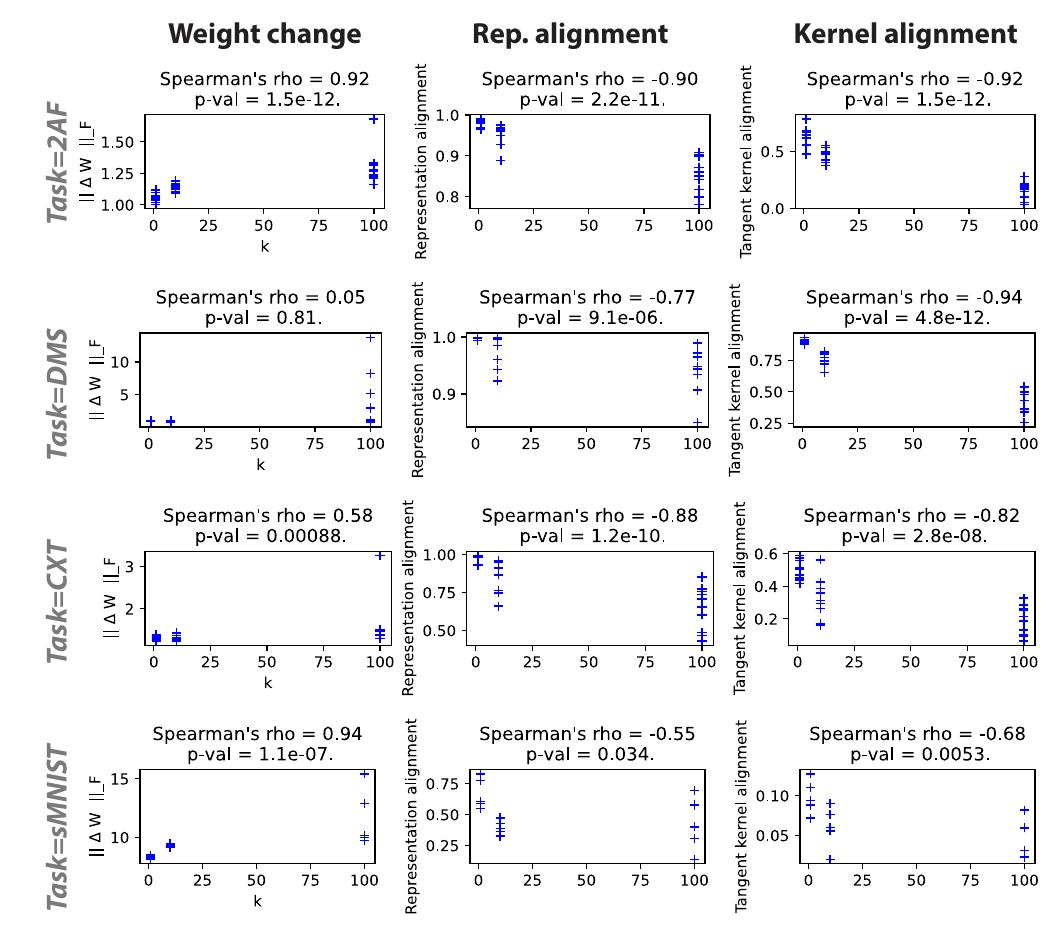}
    \caption{ \textbf{Consistent trends observed in Figure~\ref{fig:svd} also for "softer" low-rank weights}. Here, instead of the "hard" low-rank weights in Figure~\ref{fig:svd} --- where the $i^{th}$ weight singular value $s_i$ is set to $0$ if $i > r$ for rank $r$ --- we introduce a smoother decay in singular value, where we replace the singular values with $s_i = s_1 (1-i/N)^k $ after performing SVD; this means that greater $k$ leads to lower effective rank. Plotting conventions follow that of Figure~\ref{fig:svd}
    } 
    \label{fig:soft_rank}
\end{figure}

\begin{figure}[h!]
    \centering
    \includegraphics[width=0.99\textwidth]{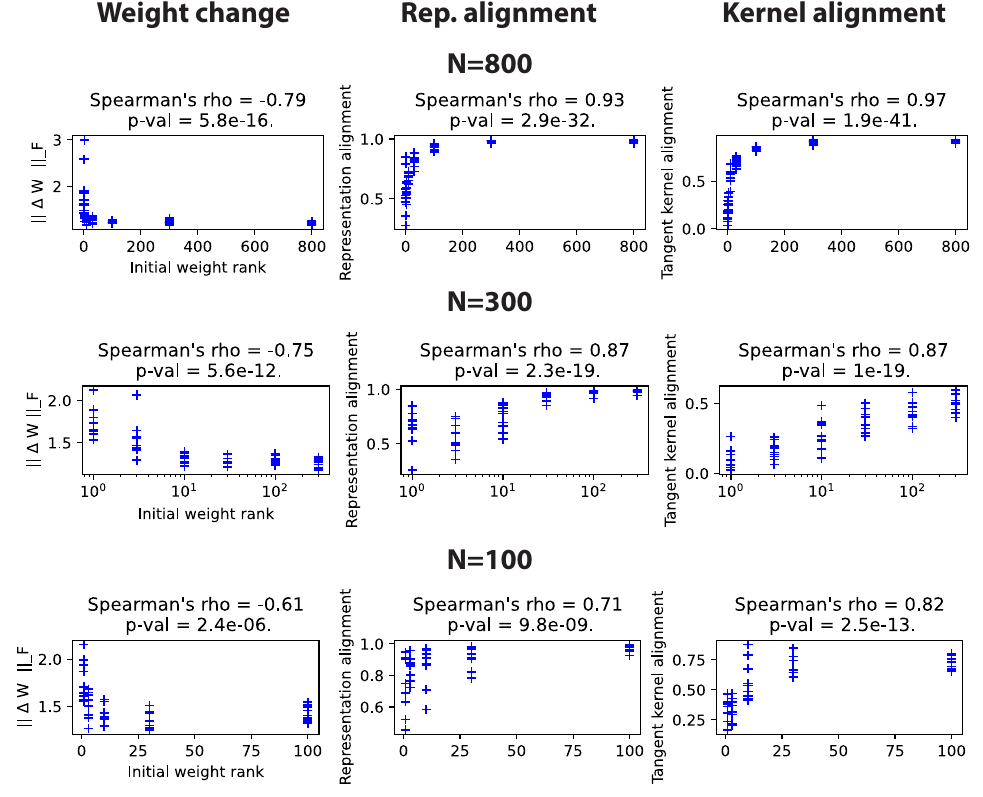}
    \caption{ \textbf{Consistent trends observed in Figure~\ref{fig:svd} across various network sizes ($N$)}. We replicated the results of Figure~\ref{fig:svd} for different values of $N$, using the CXT task as an illustrative example. However, the observed trend remains consistent for both the 2AF and DMS tasks. Plotting conventions follow that of Figure~\ref{fig:svd}. 
    } 
    \label{fig:svd_varN}
\end{figure}

\begin{figure}[h!]
    \centering
    \includegraphics[width=0.99\textwidth]{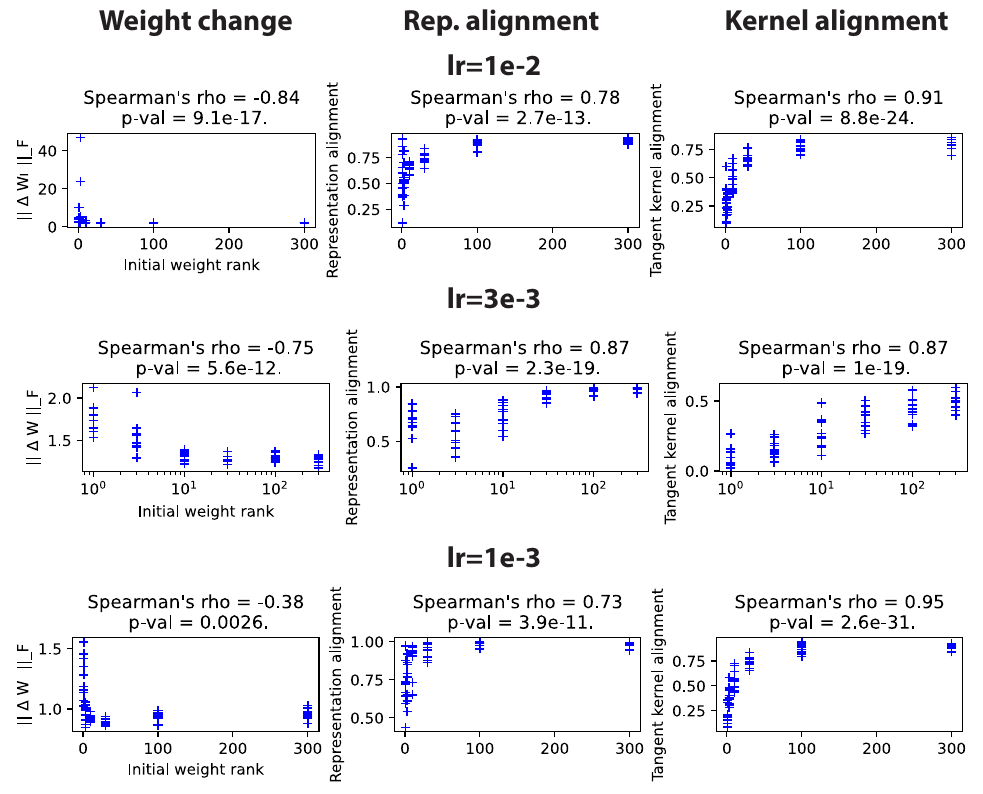}
    \caption{ \textbf{Consistent trends observed in Figure~\ref{fig:svd} across various learning rates (lr)}. We replicated the results of Figure~\ref{fig:svd} for different learning rates, using the CXT task as an illustrative example. However, the observed trend remains consistent for both the 2AF and DMS tasks. Plotting conventions follow that of Figure~\ref{fig:svd}. 
    } 
    \label{fig:svd_varLR}
\end{figure}

\begin{figure}[h!]
    \centering
    \includegraphics[width=0.99\textwidth]{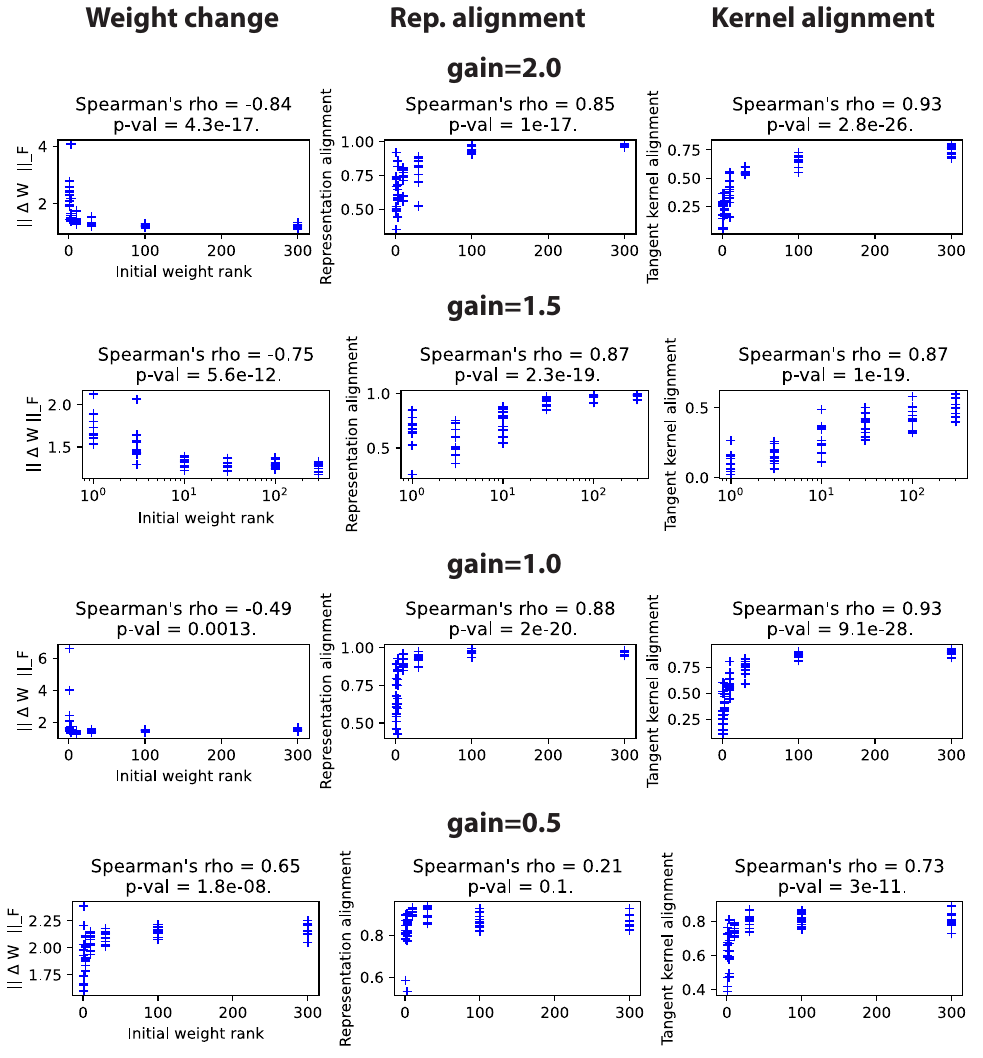}
    \caption{ \textbf{Consistent trends observed in Figure~\ref{fig:svd} across various initial gain}. Here, the gain refers to $g$, as weights are initialized as \(W_{ij} \sim \mathcal{N}(0, g^2/N)\). The trends hold for most typical range of $g$ from 1.0 to 2.0, but gets weakened for smaller values, $g < 1.0$ (a closer examination of the regime bias in such setting in RNNs is left for future work). We replicated the results of Figure~\ref{fig:svd} for different learning rates, using the CXT task as an illustrative example. However, the observed trend remains consistent for both the 2AF and DMS tasks. Plotting conventions follow that of Figure~\ref{fig:svd}. 
    } 
    \label{fig:svd_varWsig}
\end{figure}

\begin{figure}[h!]
    \centering
    \includegraphics[width=0.99\textwidth]{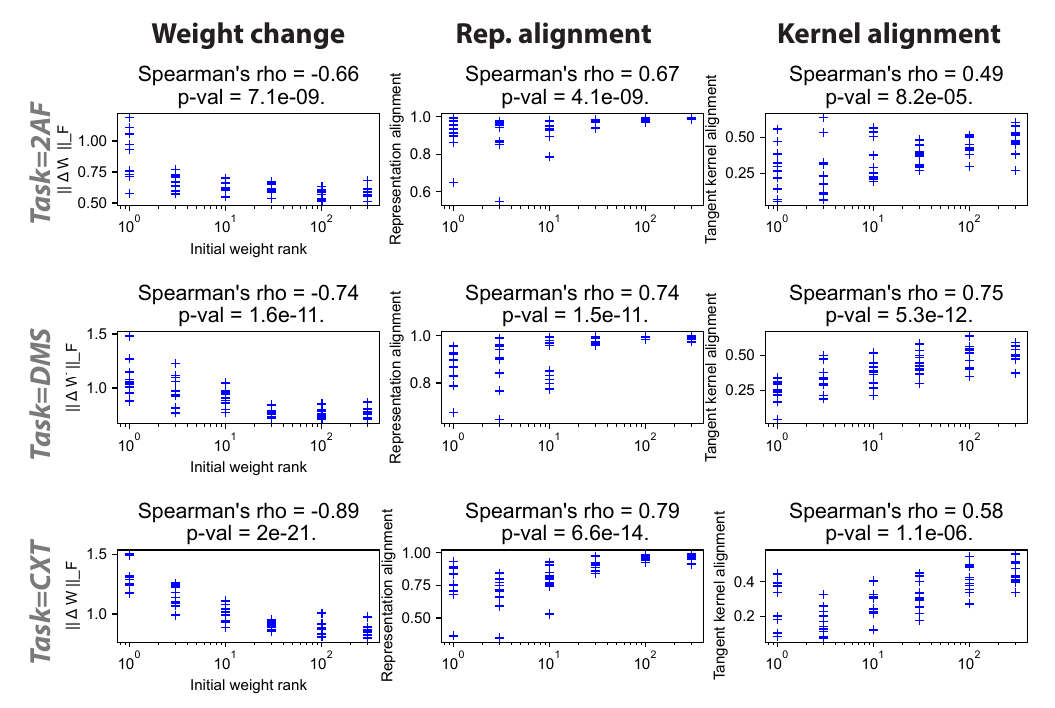}
    \caption{ \textbf{Trends in Figure~\ref{fig:svd} are also observed In training RNNs with a fivefold finer time step ($dt$) and a sequence length extended by five times}. As expected, higher rank initializations led to a marked increase in effective laziness. Plotting conventions follows that of Figure~\ref{fig:svd}. 
    } 
    \label{fig:tau25_vs_laziness}
\end{figure}

\begin{figure}[h!]
    \centering
    \includegraphics[width=0.99\textwidth]{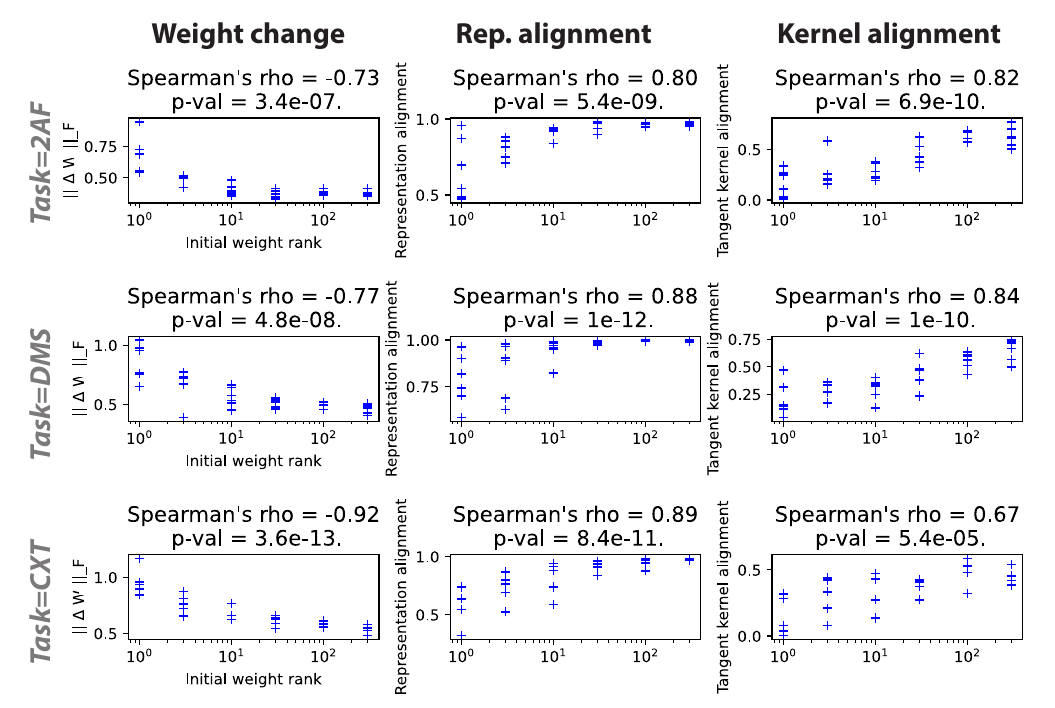}
    \caption{ \textbf{Trends in Figure~\ref{fig:svd} are also observed when fixing the leading weight eigenvalue instead of the Frobenius norm across comparisons}. As expected, higher rank initializations lead to effectively lazier learning. Plotting conventions follows that of Figure~\ref{fig:svd}. 
    } 
    \label{fig:DomEig_vs_laziness}
\end{figure}

\begin{figure}[h!]
    \centering
    \includegraphics[width=0.99\textwidth]{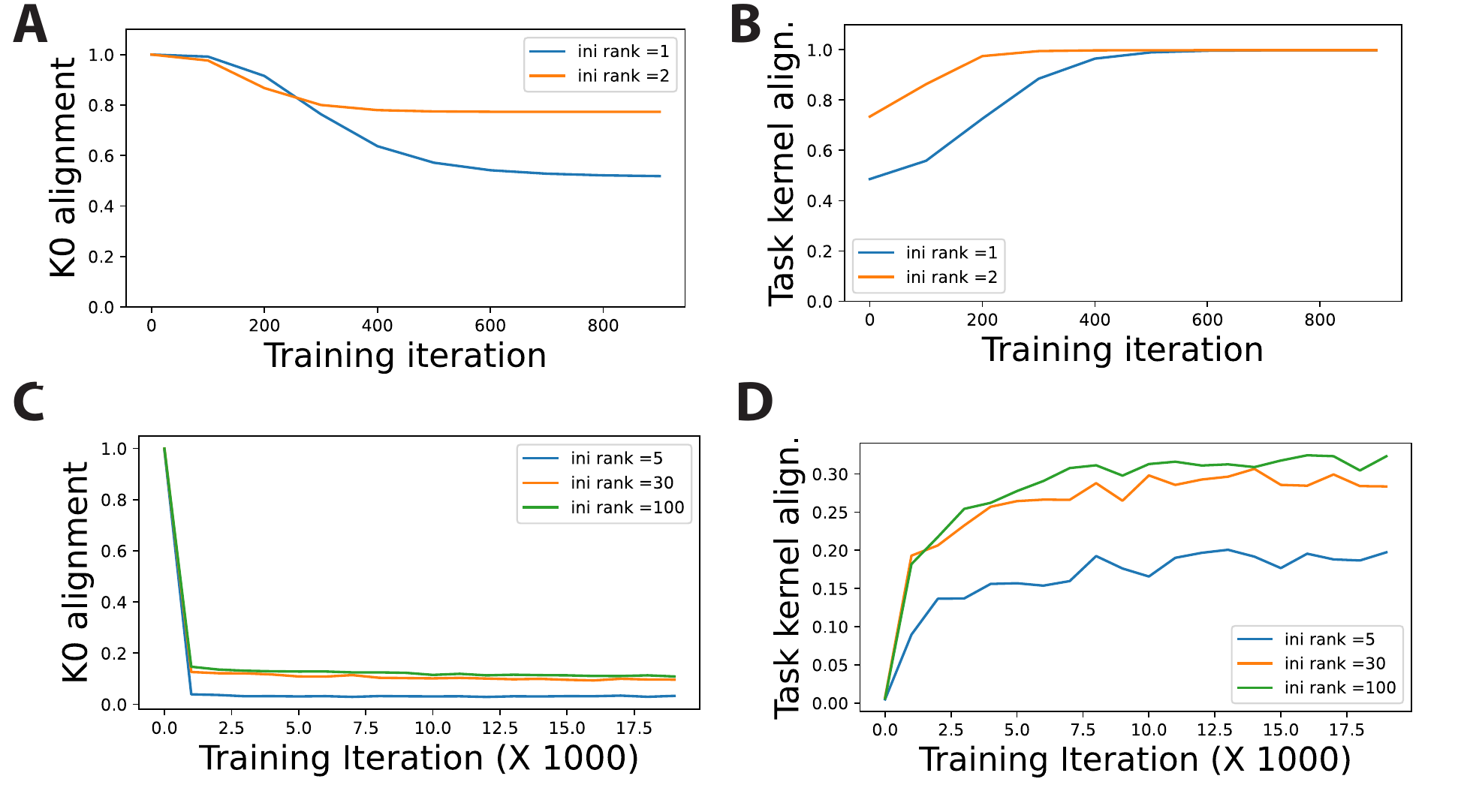}
    \caption{ \textbf{[A-B] The idealized two-layer linear network setting from Fig. 2 in \citet{atanasov2021neural}.} A) Examining the K0 alignment --- the alignment between the kernel at various training iterations and the initial kernel --- reveals that low-rank random initialization leads to greater changes during training; here, different curves correspond to different initial weight ranks. B) Despite these greater changes, networks with low-rank random initialization take longer to align with the task, as shown by the task kernel alignment metric $y^T K y / |y|^2 TrK$ throughout training. We remind the reader that $y$ corresponds to the target output and $K$ corresponds to the NTK. \textbf{[C-D] A non-idealized setting: the sMNIST task.} C) This panel shows similar trends to A). D) Similar to B), lower-rank random initializations do not achieve as high task kernel alignment within the trained iterations. This is measured by the centered kernel alignment (CKA), which assesses the kernel's alignment with class labels (Eq. 7 in \citet{baratin2021implicit}). Although higher CKA values during training could suggest enhanced feature learning (characteristic of the standard rich regime), this aligns with our findings on the effective learning regime, which focuses on changes post-training (see Introduction). Our theory in Section \ref{scn:theory} suggests that lower-rank initializations require greater changes to align with the task, which would typically require more training iterations, as seen in panel B. If training is halted prematurely, perhaps due to resource constraints (as in panel D), these initializations may achieve lower final alignment within the training period. It remains unclear if extended training would lead to similar final alignment across different initializations in a wide range of scenarios. Future research should further investigate the relationship between rankedness of initializations and their impact on the converged solution's representation, including task kernel alignment, across diverse settings.
    } 
    \label{fig:Alignment_over_training}
\end{figure}

\begin{figure}[h!]
    \centering
    \includegraphics[width=0.99\textwidth]{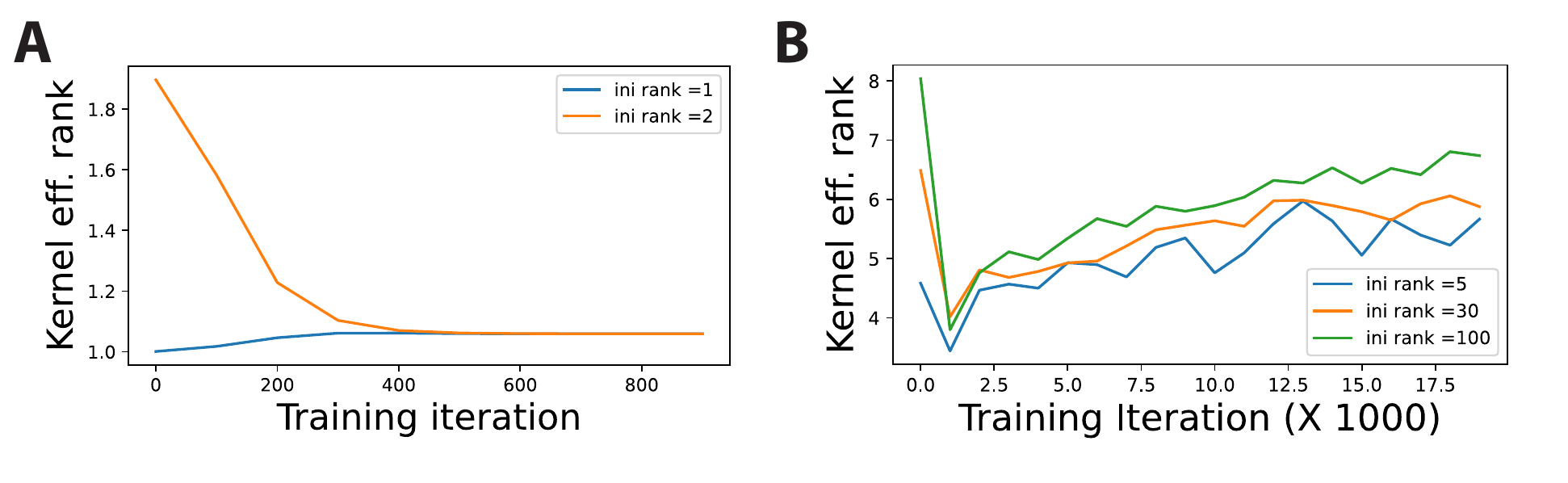}
    \caption{ The evolution of the leading NTK eigenvalue relative to the rest of the eigenvalues was tracked using an effective rank measure. This measure is based on the ratio of the kernel trace to the kernel dominant eigenvalue, i.e., $\sum_i \lambda_i / \lambda_1$, which indicates the number of eigenvalues on the order of the dominant one. We apply this analysis to A) the idealized setting and B) the sMNIST task, as used in Appendix Figure~\ref{fig:Alignment_over_training} and Figure~\ref{fig:svd}, respectively. These results suggest that the kernel effective rank approaches that of the task throughout the training process.
    } 
    \label{fig:EffRank_over_training}
\end{figure}

\begin{figure}[h!]
    \centering
    \includegraphics[width=0.99\textwidth]{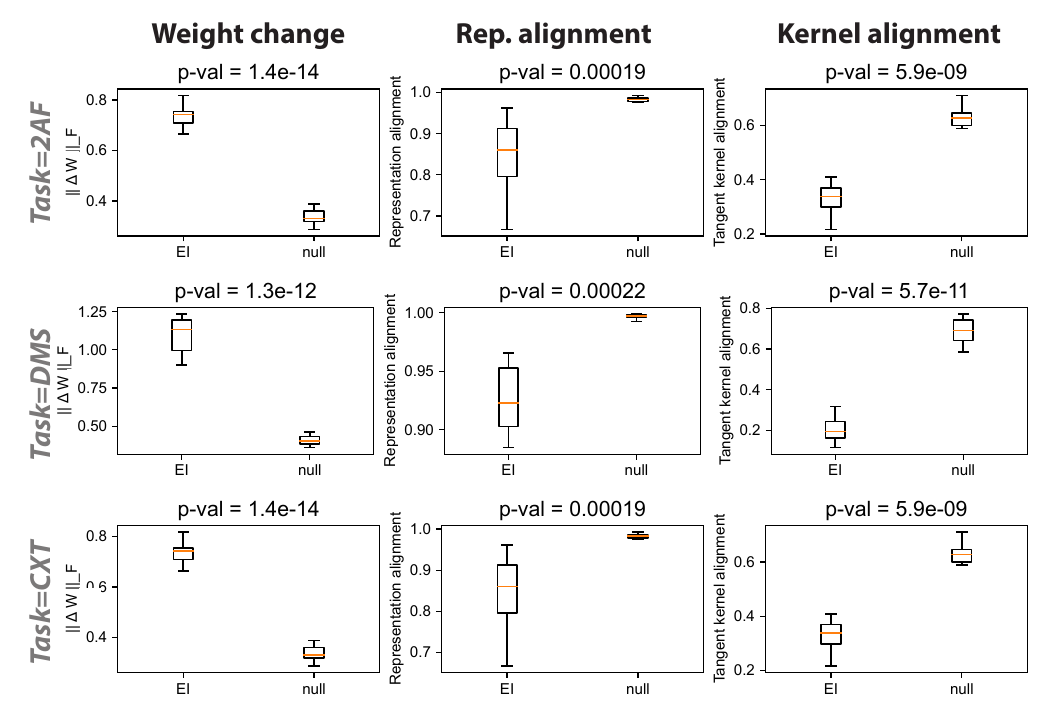}
    \caption{ Maintaining the constraint of Dale's Law during the entire training process, rather than just at initialization, produced a trend analogous to that observed in Figure~\ref{fig:bio_spectrum_laziness}. Plotting conventions follow that of Figure~\ref{fig:bio_spectrum_laziness}. 
    } 
    \label{fig:ConstrainEI_vs_laziness}
\end{figure}

\begin{figure}[h!]
    \centering
    \includegraphics[width=0.99\textwidth]{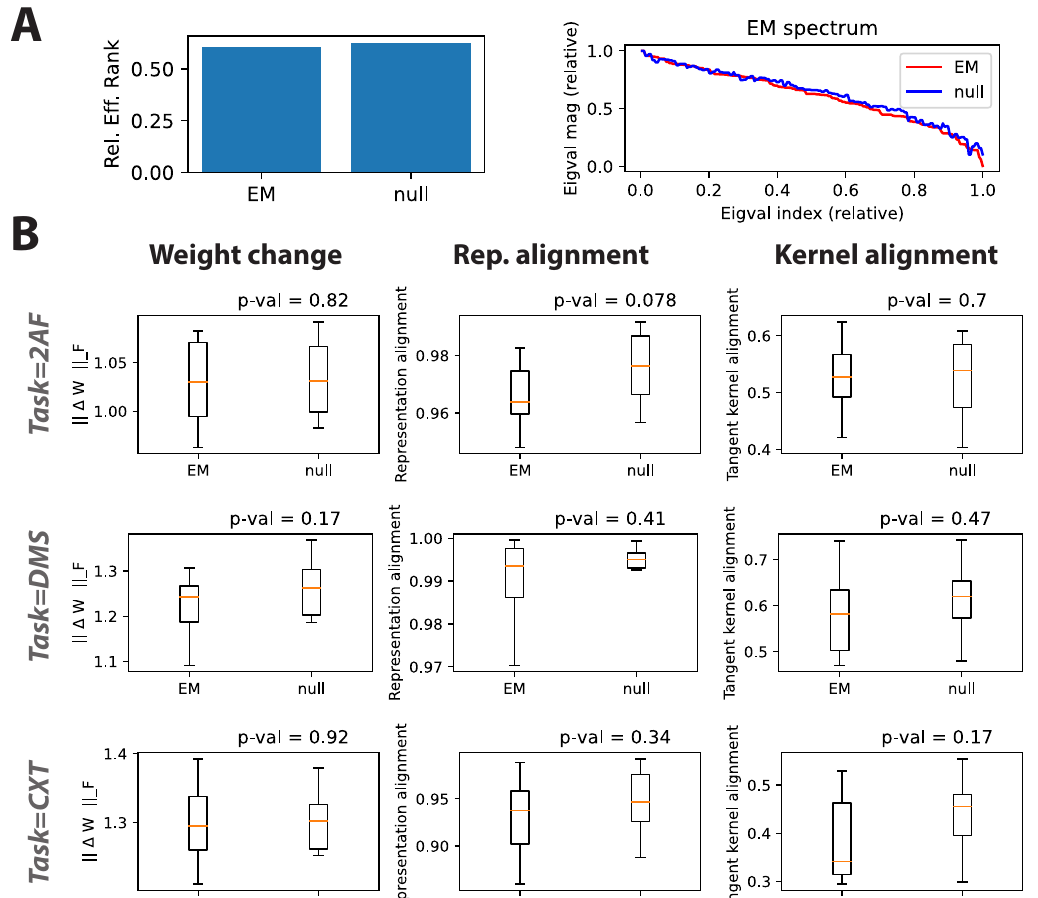}
    \caption{ \textbf{Shuffling the EM connectivity, while maintaining the sparsity structure, destroys the low-rankedness and the impact on effective laziness}. We repeated the analyses with the EM initial connectivity in Figures~\ref{fig:bio_spectrum_laziness} but performed random shuffling on the EM connectivity, to see if the low-rankedness and the impact on effective laziness is due to the sparsity in the dataset. Performing such shuffling destroys these trends. Plotting conventions follow that of Figure~\ref{fig:bio_spectrum_laziness}
    } 
    \label{fig:ShuffleEM}
\end{figure}

We perform additional simulations to show the robustness of our main trends, Low-rank initial recurrent weights lead to greater changes (or effectively richer learning) in RNNs. We show the main trends observed in Figure~\ref{fig:svd} holds also for Uniform initialization (Figure~\ref{fig:svd_uniform}), soft initial weight rank (Figure~\ref{fig:soft_rank}), various network sizes (Figure~\ref{fig:svd_varN}), learning rates (Figure~\ref{fig:svd_varLR}), initial weight gains (Figure~\ref{fig:svd_varWsig}) and Dale's Law constraint throughout training (Figure~\ref{fig:ConstrainEI_vs_laziness}), finer simulation time step $dt$ (Figure~\ref{fig:tau25_vs_laziness}) and fixing the leading initial weight eigenvalue (Figure~\ref{fig:DomEig_vs_laziness}). The trends in Figure~\ref{fig:bio_spectrum_laziness} also applies to the DMS task (Figure~\ref{fig:bio_laziness_DMS}) and the CXT task (Figure~\ref{fig:bio_laziness_CXT}). Also, without the low-rankedness in the shuffled EM connectivity, the impact on effective laziness also goes away (Figure~\ref{fig:ShuffleEM}). In Figure~\ref{fig:linear2layer} we confirm that the results, shown in Figure~\ref{fig:svd} and predicted by Theorem~\ref{thm:NTKalign}, are also observed in a two-layer linear network setup. Again, we find that in situations where initializations are random, higher rank initialization leads to greater tangent kernel alignment than lower rank cases. We have also tracked the evolution of kernel task alignment (Figure~\ref{fig:Alignment_over_training}) and kernel effective rank over the course of training (Figure~\ref{fig:EffRank_over_training}). 

\newpage 

\newpage 

\newpage 

\newpage

\newpage 

\newpage 

\newpage 

\newpage 

\newpage 

\newpage 

\newpage 

\newpage 

\begin{figure}[h!]
    \centering
    \includegraphics[width=0.8\textwidth]{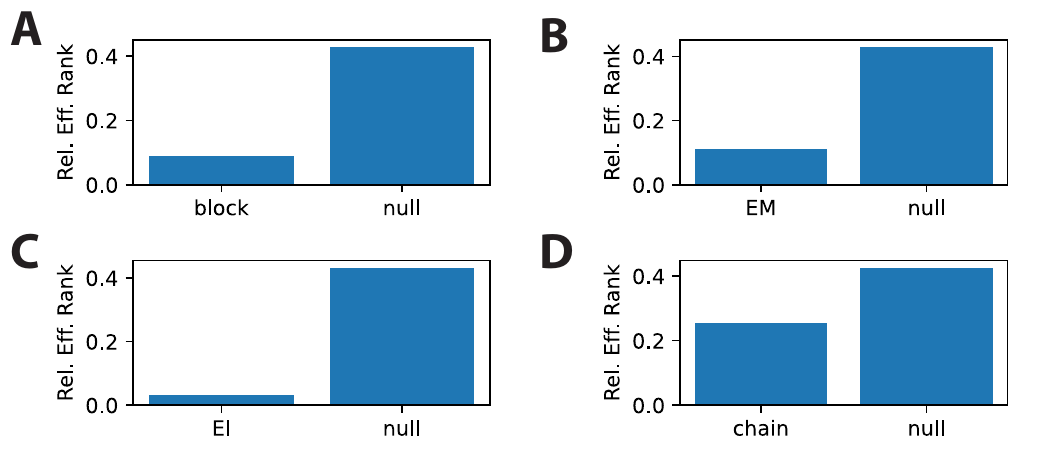}
    \caption{ Measuring connectivity effective rank based on singular values instead of eigenvalues led to a similar conclusion as Figure~\ref{fig:bio_spectrum_laziness}: these experimentally-driven connectivity structures exhibit lower effective rank compared to random Gaussian initialization (null). The plotting conventions used here follow those in Figure~\ref{fig:bio_spectrum_laziness}, with panels A-D corresponding to the ones in that figure. 
    } 
    \label{fig:Bio_spectrum_sval}
\end{figure}

\newpage 

\newpage 

\begin{figure}[h!]
    \centering
    \includegraphics[width=0.99\textwidth]{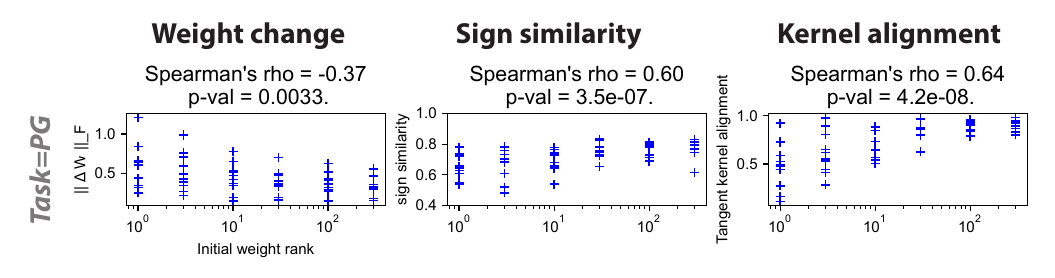}
    \caption{ Training RNNs on the pattern generation task, as illustrated in Fig. S7 of \citet{bellec2020solution}, showed consistent trends with our conclusion: initializations with higher ranks resulted in a more pronounced tendency towards effectively lazier learning. Plotting conventions follows that of Figure~\ref{fig:svd}. 
    } 
    \label{fig:PG_laziness}
\end{figure}

\newpage 

\end{document}